\documentclass[10pt,twocolumn,letterpaper]{article}

\usepackage{cvpr}              % To produce the CAMERA-READY version
\definecolor{cvprblue}{rgb}{0.21,0.49,0.74}
\usepackage[pagebackref,breaklinks,colorlinks,allcolors=cvprblue]{hyperref}

\usepackage{multirow}
\usepackage{enumitem}
\usepackage{makecell}
\usepackage{adjustbox}

\def\PaperTitle {Unsupervised Monocular 3D Keypoint Discovery \\ from Multi-View Diffusion Priors}
\def\PaperNumber {*****}
\newcommand{\Tref}[1]{Table~\ref{#1}}

\newcommand{\fref}[1]{Fig.~\ref{#1}}
\newcommand{\Fref}[1]{Figure~\ref{#1}}

\newcommand{\textblock}[1]{\noindent\textbf{#1}}

\def\eg{\emph{e.g.}}

\def\etal{\emph{et al.}\ }

\def\paperID{\PaperNumber} % *** Enter the Paper ID here
\def\confName{CVPR}
\def\confYear{2026}

\title{\PaperTitle}

\author{Subin Jeon \quad In Cho \quad Junyoung Hong \quad Woong Oh Cho \quad Seon Joo Kim\\[1mm]
Yonsei University\\
{\tt\small\{subinjeon, join, junyounghong, wocho, seonjookim\}@yonsei.ac.kr}
}

\begin{document}
% \maketitle
\twocolumn[{%
\renewcommand\twocolumn[1][]{#1}%
\maketitle
% \begin{figure}[]
% \begin{center}
% \includegraphics[width=1\linewidth]{Main/Assets/teaser.pdf}
% \end{center}
%    \caption{\textbf{KeyDiff3D} enables 3D keypoint prediction and object manipulation from a single image using multi-view diffusion priors. It generalizes effectively to in-the-wild and out-of-domain scenarios across diverse categories, including both human and animal domains.}
%    \label{fig:teaser}
%    \vspace{-10pt}
% \end{figure}

\begin{center}
    \centering
    \includegraphics[width=\textwidth]{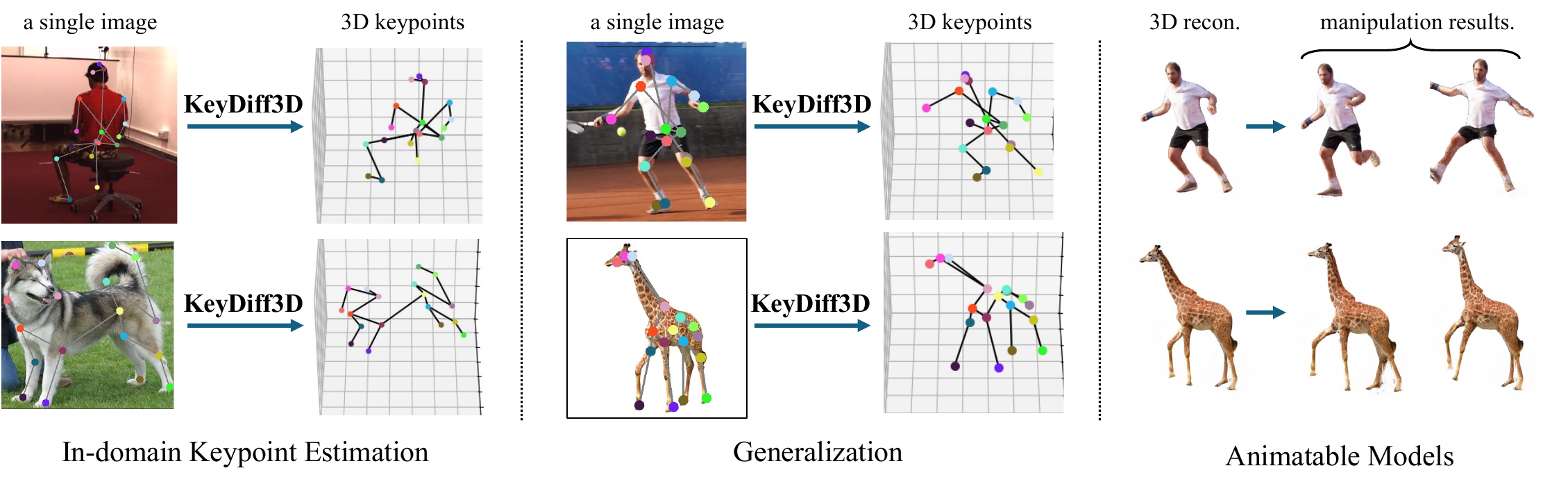}
    \vspace{-20pt} 
    \captionof{figure}{
        \textbf{KeyDiff3D} enables 3D keypoint prediction and object manipulation from a single image using multi-view diffusion priors. It generalizes effectively to in-the-wild and out-of-domain scenarios across diverse categories, including both human and animal domains.}
    \label{fig:teaser}
    \vspace{-1pt}
\end{center}
}]

\begin{abstract}
Most existing 3D keypoint estimation methods rely on manual annotations or calibrated multi-view images, both of which are expensive to collect.
This paper introduces KeyDiff3D, a framework that can accurately predict 3D keypoints from a single image, thus eliminating the need for such expensive data acquisitions.
To achieve this, we leverage powerful geometric priors embedded in a pretrained multi-view diffusion model.
In our framework, the diffusion model generates multi-view images from a single image, serving as supervision signals to provide 3D geometric cues to our model.
We also introduce a 3D feature extractor that transforms implicit 3D priors embedded in the diffusion features into explicit 3D feature volumes.
Beyond accurate keypoint estimation, we further introduce a pipeline that enables manipulation of 3D objects generated by the diffusion model.
Experimental results on diverse datasets, including Human3.6M, CUB-200-2011, Stanford Dogs, and several in-the-wild and out-of-domain inputs, highlight the effectiveness of our method in terms of accuracy, generalization, and its ability to enable manipulation of 3D objects generated by the diffusion model from a single image.
\end{abstract}

\section{Introduction}
Understanding the 3D structure of objects is fundamental for various downstream tasks, including pose estimation, animation, and interaction analysis.
% Discovering 3D structures is essential for understanding 3D objects, enabling a wide range of applications such as pose estimation, animation, and interaction understanding.
A common approach is to represent structures using a set of 3D keypoints -- such as body joints or facial landmarks -- that provide a compact and interpretable abstraction of an object's geometry.
Learning such representations, however, typically requires thousands of 3D annotations, which are expensive to collect and thus largely limited to well-studied domains like humans.

On the other hand, discovering 3D keypoints in an unsupervised manner \cite{suwajanakorn2018keypointnet,sun2023bkind3d,honari2024unsupervised} offers a promising alternative.
By removing the need for manual annotations, this approach can be easily scaled to generic or rare object categories.
Most existing methods accomplish this by training neural networks via image reconstruction, where multi-view images are used as inputs or reconstruction targets.
Although this scheme alleviates the cost for manual annotations, obtaining calibrated, object-centric multi-view images still requires a controlled capture environment.
This limits the diversity of multi-view datasets, making it difficult to extend these methods to diverse and in-the-wild inputs.

Identifying 3D keypoints from monocular images can be a key for scalable 3D understanding, as monocular images are far more accessible than multi-view images.
Yet, recovering 3D structures from a single image -- without camera parameters or manual annotations -- is a fundamentally under-constrained problem, posing challenges arising from severe depth ambiguity and occlusion.
Despite its potential, such challenges have left monocular 3D keypoint estimation largely unexplored.

In this paper, we present a simple yet effective framework for unsupervised monocular 3D keypoints estimation.
Our approach accurately predicts 3D keypoints from a single image, can be trained only using unconstrained single-view images, and generalizes well to in-the-wild and out-of-domain inputs.
It is achieved by leveraging powerful geometric priors embedded in pretrained multi-view diffusion models, throughout both our training pipeline and model design.
For training, the diffusion model is used to synthesize multi-view images that serve as supervision signals.
By learning to reconstruct these generated images, our model no longer requires expensive data acquisitions -- 3D annotations, camera parameters, or controlled capturing environments -- making it scalable to diverse object categories.

The core of our model design lies in employing the diffusion model as a 2D multi-view feature extractor.
Instead of directly using generated multi-view images as inputs, we propose to extract intermediate representations from the multi-view diffusion model.
Specifically, our model extracts representations from multiple layers of the diffusion model and fuses them to form rich multi-view features.
The extracted multi-view features are unprojected into 3D space to construct volumetric representations, from which we estimate 3D keypoints and their dense connectivity graphs.
This scheme allows our model to exploit the diffusion model's inherent 3D priors directly, rather than relying on noisy generated images, leading to more accurate and robust 3D keypoint estimation from a single image.

In addition to accurate 3D keypoints estimation, our model -- built upon the diffusion model -- produces 3D keypoints that are inherently aligned with the coordinate systems learned by the diffusion backbone.
Building on this property, we introduce a pipeline that enables manipulation of 3D objects generated by the diffusion model.
Concretely, we first reconstruct 3D objects from generated multi-view images. Then we extract directed tree-structured edges based on the estimated keypoints and connectivity graphs, and compute skinning weights with a Gaussian-based distance function.
This yields 3D objects that can be easily manipulated in a way similar to commonly used skeletons.

% %%% [in] cite 넣어야 할까? MST extraction, Gaussian-based distance
% % changed: Furthermore, our model -- built upon the diffusion model -- produces 3D features that are inherently aligned with the coordinate systems learned by the diffusion backbone.
% \joinrev{Since our model is built upon the diffusion model, it produces 3D features that are in inherently aligned with the coordinate systems learned by the diffusion backbone.}
% % by the diffusion model -> from a single image
% Building on this property, we introduce a pipeline that enables manipulation of 3D objects generated \joinrev{from a single image}.
% \joinrev{We first reconstruct 3D objects from generated multi-view images, and then extract directed tree-structured edges based on the estimated keypoints and connectivity graphs.
% Skinning weights are computed with a Gaussian-based distance function, yielding 3D objects that can be easily manipulated in a way similar to commonly used skeletons.}
% % We first reconstruct 3D objects from generated multi-view images. Then we extract directed tree-structured edges based on the estimated keypoints and connectivity graphs, and compute skinning weights with a Gaussian-based distance function.
% % This yields 3D objects that can be easily manipulated in a way similar to commonly used skeletons.

We validate our method, dubbed Keypoints with Diffusion in 3D (KeyDiff3D), on diverse aspects and datasets.
These include Human3.6M \cite{ionescu2013h36m}, CUB-200-2011 \cite{wah2011cub}, and Stanford Dogs \cite{stanford_dogs} for diverse object categories, as well as several in-the-wild and out-of-domain images.
Experimental results highlight the effectiveness and the versatility of our method, demonstrating that our method accurately estimates 3D keypoints and can be easily scaled to diverse object categories.
We also showcase several animatable 3D objects obtained using the diffusion model and our method.
\section{Related work}
\textblock{Monocular 3D pose estimation.} 3D pose estimation has been studied across both human~\cite{sarafianos20163d} and animal domains~\cite{pierre2021anipose, ruegg2023bite, gunel2019deepfly3d}, with existing approaches broadly categorized into multi-view and monocular settings based on input modality.
While multi-view methods achieve high accuracy by leveraging geometric constraints such as triangulation~\cite{zhao2023triangulation, iskakov2019learnable, liao2024multiple}, bundle adjustment~\cite{usman2022metapose}, or unprojection-based volumetric aggregation~\cite{pavlakos2017harvesting, burenius20133d}, they typically rely on synchronized camera setups, which may limit their scalability in unconstrained environments.

Monocular 3D pose estimation, in contrast, has been widely studied for its practicality in real-world scenarios, yet it remains fundamentally ill-posed due to depth ambiguities and occlusions.
Supervised methods \cite{li20143dhpe, park20163d, martinez2017simple, zhou2017towards} address this by training on large-scale datasets with ground-truth 3D annotations, but their reliance on annotations and constrained capture environments limits their scalability.  
To reduce 3D annotation costs, a variety of unsupervised methods have been proposed, utilizing 2D keypoint annotations \cite{chen2019gss, kocabas2019self, iqbal2020weakly, li2020geometry, wandt2021canonpose, drover2018can, wandt2022elepose, sosa2023self}, unpaired 3D data \cite{wandt2019repnet, kundu2020pgnis}, or predefined structural priors or templates \cite{kundu2020kinematic, yang2024mask}.
Although these priors enable strong performance even in monocular settings, their human-specific nature limits generalization, leaving most non-human categories largely unexplored.

\textblock{Unsupervised keypoint discovery.}
Unsupervised keypoint discovery aims to learn meaningful structural representations without any manual annotation.  
A common strategy in 2D settings is to use transformation-aware consistency as a self-supervised signal, enforcing keypoints to remain coherent under various transformations. 
Equivariance-based approaches~\cite{Thewlis_2017_ICCV, hedlin2024stable_keypoints} achieve this by training keypoint detectors whose predictions transform consistently under synthetic transformations.
Alternatively, several methods~\cite{siarohin2019animating, siarohin2021co_part, siarohin2019fomm, kulkarni2019unsupervised, minderer2019unsupervised_dynamics, kim2019unsupervised, sun2022bkind} leverage natural variations across time or viewpoints, encouraging keypoints to capture motion- or pose-dependent changes.
Both paradigms are often combined with reconstruction objectives \cite{jakab2018unsupervised, zhang2018unsupervised, lorenz2019part_disentangling, he2022autolink}, where models learn to reconstruct images using the predicted keypoints and corresponding appearance features.

Several methods \cite{suwajanakorn2018keypointnet, sun2023bkind3d, honari2024unsupervised} extend unsupervised keypoint discovery to 3D using multi-view data with known camera parameters.
KeypointNet \cite{suwajanakorn2018keypointnet} learns to predict 3D keypoints from a single image by enforcing multi-view consistency during training.
BKinD-3D \cite{sun2023bkind3d} learns 3D volumetric heatmaps from calibrated images by reconstructing multi-view video sequences. %, requiring at least two calibrated views per frame at both training and inference.
Honari \etal~\cite{honari2024unsupervised} triangulate 3D keypoints from multi-view self-supervised 2D predictions and distill the model into a single-view 2D-to-3D lifting network.
Although some of these methods enable single-image inference, they still depend on multi-view calibration or supervision during training, limiting their scalability to settings where only monocular data is available.

% \textblock{Multi-view diffusion models. }
% Image diffusion models have been extended to generate multi-view images from text~\cite{shi2023mvdream} or image~\cite{liu2023zero123, liu2024syncdreamer} inputs.
% Building on the 2D prior of pretrained models, these methods are fine-tuned on synthetic or multi-view datasets to enable view conditioning and enforce cross-view consistency.
% % we use mv diffuion model as our powerful 3d representation

\textblock{Pretrained diffusion models for visual understanding.}
Pretrained diffusion models have recently emerged as strong visual priors for a wide range of perception tasks.
They learn semantically meaningful features that can be directly exploited for downstream applications, \eg, semantic segmentation \cite{baranchuk2021label, xu2023open, namekata2024emerdiff} and correspondence matching \cite{luo2023hyperfeats, hedlin2023unsupervised, tang2023emergent, zhang2023tale}, and representation learning \cite{yang2023diffusionrl, yue2024exploring}.
% Features extracted from these models have also been used for representation learning \cite{yang2023diffusionrl, yue2024exploring}, showing strong transferability across tasks.
Most relevant to our work, StableKeypoints \cite{hedlin2024stable_keypoints} leverages a 2D diffusion model for unsupervised 2D keypoint detection.
We extend this paradigm to 3D by utilizing the spatial priors embedded in a multi-view diffusion model for unsupervised monocular 3D keypoint estimation.

\begin{figure*}[t]
\centering
\includegraphics[trim={0 20 0 25pt}, width=1\linewidth]{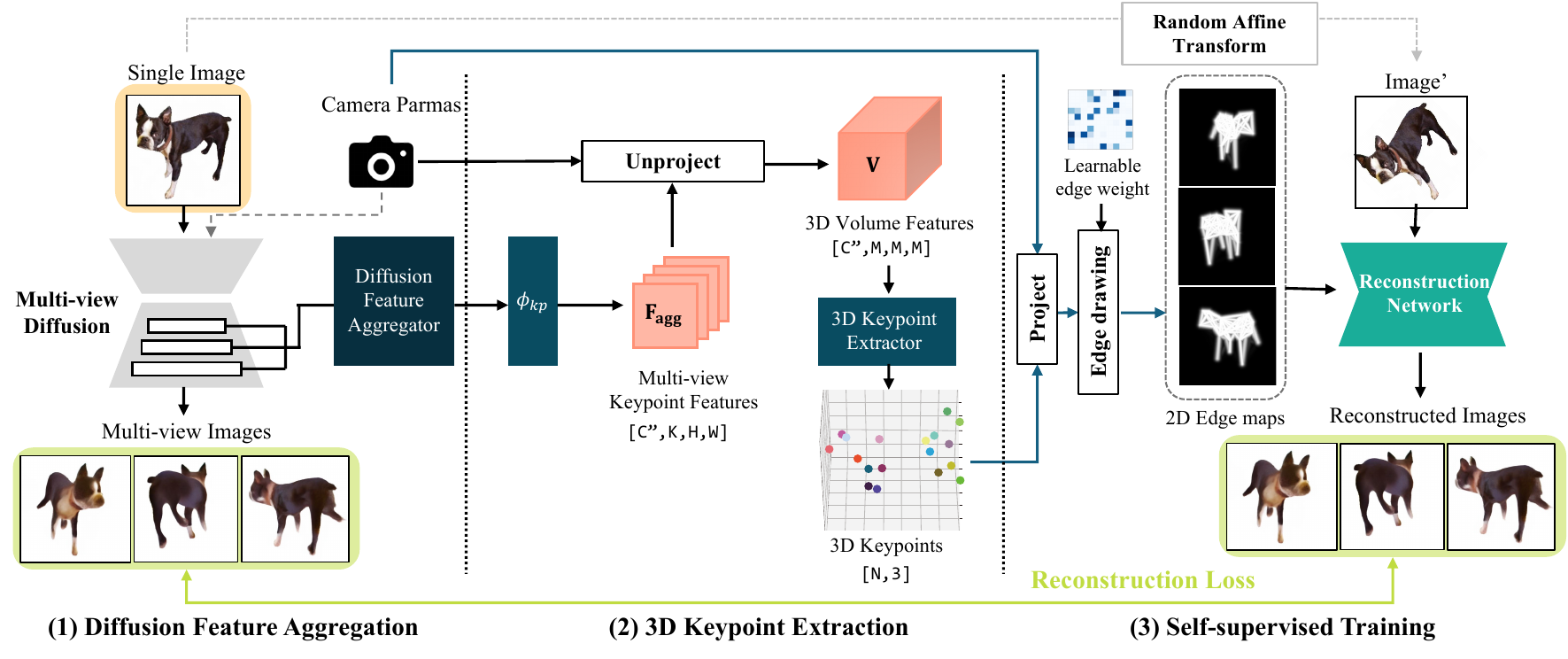}
\vspace{-5pt}
\caption{The overall pipeline of KeyDiff3D. From a single image, \textbf{(1)} a pretrained multi-view diffusion model provides novel views and multi-view features, \textbf{(2)} which are aggregated and lifted into a 3D feature volume for keypoint prediction, and \textbf{(3)} the predicted 3D keypoints are projected to the generated views to provide structural cues for self-supervised reconstruction}
\label{fig:overview}
\vspace{-10pt}
\end{figure*}

\section{Method}
We present an unsupervised framework for estimating 3D keypoints and their structural connectivity from a single image across arbitrary object categories.
Given an input image $I$, our method predicts a set of 3D keypoints $\mathbf{S}=\{\mathbf{s}_n\}_{n=1}^{N}$ and a learnable adjacency matrix $\mathcal{A} \in \mathbb{R}^{N \times N}$, where each entry $e_{i,j}$ denotes the learnable edge weight between keypoints $i$ and $j$.
Here, $\mathbf{s}_n \in \mathbb{R}^3$ represents the predicted 3D location of the $n$-th keypoint.
% Given an input image $I$, our method predicts a set of 3D keypoints $\mathbf{S}=\{\mathbf{s_n}\}_{n=1}^{N}$ and a learnable adjacency matrix $ \mathcal{A} = \{e_{i,j} \mid i, j \in [1, N] \} $.
% Here, $\mathbf{s}_n \in \mathbb{R}^3$ represents the predicted 3D location of the $n$-th keypoint, and $e_{i,j}$ denotes a learnable weight between the keypoint pair $(i,j)$.
The key idea is to exploit geometric priors embedded in pretrained multi-view diffusion models.
By using their generated multi-view images as a supervision signal and employing diffusion models as a multi-view feature extractor, we transform implicit 3D priors in their representations into explicit 3D feature volumes.
% The key idea is to exploit geometric priors embedded in pretrained multi-view diffusion models by using generated multi-view images as a supervision signal, and transforming implicit 3D priors in their internal representations into explicit 3D feature volumes.

\Fref{fig:overview} illustrates the overview of our method. Our method comprises three main components: (1) \textit{Diffusion Feature Aggregation}, where multi-view diffusion features are extracted from a single image; (2) \textit{3D Keypoint Extraction}, which lifts multi-view features to 3D volume features and estimates the 3D keypoints; and (3) \textit{Self-Supervised Training Pipeline}, employing multi-view outputs of the diffusion model to provide self-supervised signals for learning 3D structural representations.

\subsection{Background}
Diffusion models generate images by progressively denoising samples drawn from Gaussian noise via a learned reverse process. 
Recently, this framework has been extended to the multi-view setting, where the goal is to generate a set of geometrically consistent novel views under varying camera poses.
Multi-view diffusion models with image-based conditioning \cite{voleti2024sv3d, kong2024eschernet, liu2024syncdreamer, wang2023imagedream} generate novel views from a single input image $I$, guided by a set of target camera parameters $\{\mathbf{P}_k\}_{k=1}^{K}$.
Each $\mathbf{P}_k \in \mathbb{R}^{3 \times 4}$ denotes the full camera projection matrix, defined as $\mathbf{P}_k=\mathbf{K}[\mathbf{R}_k|\mathbf{t}_k]$, where $\mathbf{K}$ is the intrinsic matrix and $(\mathbf{R}_k,\mathbf{t}_k)$ are the extrinsic parameters (rotation and translation) for a $k$-th view.
The input image $I$ serves as the source of both appearance and structural cues, while $\mathbf{P}_k$ defines the spatial configuration of each output view.
These models denoise a 4D latent tensor $\textbf{z}_t \in \mathbb{R}^{C\times K \times H \times W}$, where $H$, $W$ denote the spatial resolution and $C$ is the channel dimension.
At each timestep $t$, a denoising U-Net jointly predicts the noise residuals across all viewpoints, conditioned on both the input image $I$ and the corresponding camera parameters.

\subsection{Diffusion-guided 3D keypoint estimation}

\textblock{Diffusion feature aggregation. }
To extract structural features from a pretrained image-conditioned multi-view diffusion model, we run the denoising process from timestep $t=T$ (pure noise) to a target timestep $\tau$, and cache intermediate decoder features from the U-Net at a timestep $\tau$.
Inspired by Diffusion Hyperfeatures \cite{luo2023hyperfeats}, we train a lightweight feature aggregation network that fuses multi-layer features using learnable mixing weights.
Each intermediate feature map $\mathbf{f}_l$ for a layer $l$ is first upsampled to a shared spatial resolution and projected through a 
bottleneck layer $B_l$ to a unified channel dimension $C'$.
The final aggregated feature is defined by a weighted summation of features:
\begin{equation}
   \mathbf{F}_{\text{agg}} = \sum_{l=1}^{L} w_l \cdot B_l(\mathbf{f}_l), \; \; \; \; \mathbf{F}_{\text{agg}} \in \mathbb{R}^{C' \times K \times H \times W},
\end{equation}
where $w_l\in \mathbb{R}$ is a learned scalar weight for a layer $l$.
The aggregation network is trained with a self-supervised objective to extract pose-relevant features that are consistent across multiple generated views. 
This yields a compact representation containing structural priors from the diffusion model.

\textblock{Lifting from 2D to 3D. }
To estimate 3D keypoints, we lift the aggregated 2D multi-view features into a 3D volumetric representation via unprojection. 
We first apply a shallow keypoint head $\phi_{\text{kp}}$ to transform the aggregated features into multi-view keypoint features:
\begin{equation}
    \mathbf{F}_{\text{kp}}=\phi_{\text{kp}}(\mathbf{F}_{\text{agg}}), \;\;\;\; \mathbf{F}_{\text{kp}} \in \mathbb{R}^{C'' \times K \times H \times W},
\end{equation}
where $C''$ is the keypoint feature dimension.
For unprojection, we define a 3D feature volume of $M \times M \times M$ uniformly sampled voxels, defined in a canonical 3D coordinate space that is aligned with the world space learned by the diffusion model.
% This results in a 3D feature volume of $M \times M \times M$ uniformly sampled voxels, defined in a canonical 3D coordinate space that is aligned with the view space learned by the diffusion model.
For each voxel center $\mathbf{x} \in \mathbb{R}^3$, we project it into the image plane of view $k$ using $\mathbf{P}_k$, resulting in a subpixel coordinate $u_k(\mathbf{x}) \in \mathbb{R}^2$. 
We then sample the feature value at this location using bilinear interpolation:
\begin{equation}
    \mathbf{f}_k(\mathbf{x}) = \text{bilinear\_sample}(\mathbf{F}^{(k)}_{\text{kp}}, u_k(\mathbf{x})), 
\end{equation}
where $\mathbf{F}^{(k)}_{\text{kp}} \in \mathbb{R}^{C'' \times H \times W}$ is the feature map for a view $k$.
These features are aggregated across views using softmax-based attention along the view dimension:
\begin{equation}
    \mathbf{V}(\mathbf{x})=\sum_{k=1}^{K} \omega_k \cdot \mathbf{f}_k(\mathbf{x}), \;\;\;\; \omega_k = \text{softmax}_k(\{\mathbf{f}_k\}_{k=1}^K),
\end{equation}
resulting in the final volumetric feature map $\mathbf{V}\in\mathbb{R}^{C''\times M \times M \times M}$ for $N$ keypoints.
% The sampled features from all views are aggregated to construct a volumetric keypoint feature:

% where $\omega_k$ is a weight for each view computed by view dimension softmax operation.

\textblock{3D keypoint estimation. }
%%% [in] H_n shape check
We apply a 3D convolution network $\phi_{\text{vol}}$ to the volumetric feature $\textbf{V}$ to produce heatmaps $\mathbf{H}_n \in \mathbb{R}^{M \times M \times M}$ for each keypoint $n$.
To obtain 3D coordinates, we use the integral regression \cite{sun2018integral, iskakov2019learnable} approach, computing the predicted keypoint location via softmax:
\begin{equation}
    \mathbf{s}_n = \sum_{\mathbf{x} \in \Omega} \mathbf{x} \cdot \text{softmax}(\mathbf{H}_n(\mathbf{x})),
\end{equation}
where $\Omega$ denotes the set of all voxel centers, and $\mathbf{s}_n \in \mathbb{R}^3$ is the predicted 3D position of the keypoint $n$. This formulation enables fully differentiable keypoint localization.

\subsection{Training pipeline}
To supervise the learning of 3D keypoints from a single image, we leverage multi-view supervision using synthetic novel views generated by the pretrained diffusion model.

\textblock{Projection and structural representation. }
Given the predicted 3D keypoints $\mathbf{S} \in \mathbb{R}^{N \times 3}$, we project them onto $K$ diffusion-generated views using the corresponding camera projection matrices $\{ \mathbf{P}_k \}_{k=1}^K$, which can be expressed as
\begin{equation}
    \mathbf{S}^{(k)}_{\text{hom}} = \left[ \mathbf{S} \,\, \mathbf{1} \right] \cdot \mathbf{P}_k^\top,
\end{equation}
where $\mathbf{S}^{(k)}_{\text{hom}} \in \mathbb{R}^{N \times 3}$ denotes the projected keypoints in homogeneous coordinates, and the 2D keypoints $\mathbf{S}^{(k)} \in \mathbb{R}^{N \times 2}$ are obtained via homogeneous normalization.
% % \begin{equation}
% %     \mathbf{S}^{(k)}_{hom} = \left[ \mathbf{S} \,\, \mathbf{1} \right] \cdot \mathbf{P}_k^\top
% % \end{equation}
% where the result is in homogeneous coordinates, and 2D keypoints \( \mathbf{S}^{(k)} \in \mathbb{R}^{N \times 2} \) are obtained by normalizing with respect to the third coordinate.
% where \( \mathbf{S}^{(k)} \in \mathbb{R}^{N \times 2} \) denotes the projected 2D keypoints in view \( k \).
To further constrain keypoint predictions, we construct a soft edge map from the projected keypoints by drawing differentiable Gaussian lines $ \mathbf{L}_{i,j}^{(k)} \in \mathbb{R}^{H \times W} $ between each keypoint pair $ (i,j) $, as in \cite{he2022autolink, sun2023bkind3d}.
%using the learnable adjacency matrix $ \mathcal{A}$. 
% Specifically, we draw differentiable Gaussian lines $ \mathbf{L}_{i,j}^{(k)} \in \mathbb{R}^{H \times W} $ between each keypoint pair $ (i,j) $, using the learnable adjacency matrix $ \mathcal{A}$. 
Specifically, for each pixel $u$ and keypoints $(\mathbf{S}_i, \mathbf{S}_j)$, we compute the distance-to-line-segment 
$t=\mathrm{clamp}\!\left(\frac{(u-p_i)^\top v}{v^\top v},0,1\right),$
$d=\|u-(p_i+t v)\|_2,\ \text{where } v=p_j-p_i.$
We render the Gaussian line response as $\mathbf{L}_{ij}(u)=w_{ij}\exp\!\left(-d^2/(2\sigma^2)\right)$, where the learnable adjacency matrix $\mathcal{A}$ provides a multiplicative gate $w_{ij}=\mathrm{softplus}(A_{ij})$.
% Specifically, we define a learnable adjacency matrix $ \mathcal{A} = \{e_{i,j} \mid i, j \in [1, N] \} $, where $e_{i,j}$ is a learnable weight between each keypoint pair $ (i,j)$, and draw differentiable Gaussian lines $ \mathbf{L}_{i,j}^{(k)} \in \mathbb{R}^{H \times W} $. % between each connected keypoint pair \( (i,j) \in \mathcal{A} \).
The final edge map \( \mathbf{E}^{(k)} \in \mathbb{R}^{1 \times H \times W} \) is obtained by aggregating the per-pair line maps using a pixel-wise maximum operation.
These edge maps are then concatenated with the 2D keypoint heatmaps \( \mathbf{Q}^{(k)} \in \mathbb{R}^{N \times H \times W} \) and fed into the reconstruction network, providing explicit structural cues for each view.

\textblock{Multi-view reconstruction loss.}
We train a reconstruction network to synthesize each target view $k$ using an appearance feature extracted from the input image $I$ and structural cues from the projected 2D keypoints.
% We train a reconstruction network to synthesize each target view \( k \) using both the input image \( I \) and the structural representation derived from the predicted 3D keypoints.
% The network takes as input:
% (1) an appearance feature extracted from the input image, and
% (2) the edge map \( \mathbf{E}^{(k)}\), constructed from the projected 2D keypoints.
The input image is first augmented with a random affine transformation.
The network then predicts a target image $ \hat{I}^{(k)} \in \mathbb{R}^{3 \times H \times W} $, which is supervised using the diffusion-generated reference image $ I^{(k)} $.  
We use a perceptual loss computed on VGG features and a mask loss on the predicted edge map.
The overall training objective is defined as
\begin{equation}
    \mathcal{L} = \frac{1}{K} \sum_{k=1}^{K} \left( \lambda_{\text{vgg}} \cdot \mathcal{L}_{\text{vgg}}^{(k)} + \lambda_{\text{mask}} \cdot \mathcal{L}_{\text{mask}}^{(k)} \right),
\end{equation}
\begin{equation}
\mathcal{L}_{\text{vgg}}^{(k)} = \left\| \psi(\hat{I}^{(k)}) - \psi(I^{(k)}) \right\|_1, \mathcal{L}_{\text{mask}}^{(k)} = \left\| \mathbf{E}^{(k)} - \mathbf{M}^{(k)} \right\|_2^2,
\end{equation}
% \begin{equation}
%     \mathcal{L}_{\text{mask}}^{(k)} = \left\| \mathbf{E}^{(k)} - \mathbf{M}^{(k)} \right\|_2^2,
% \end{equation}
where $\psi(\cdot)$ is the VGG feature extractor and $\mathbf{M}^{(k)} \in \{0, 1\}^{H \times W}$ is the binary foreground mask extracted from \( I^{(k)} \).
Through this training scheme, our model is trained to predict 3D keypoints without expensive manual annotations or multi-view images.

%%% maybe need cites?
% RigGS: Rigging of 3D Gaussians for Modeling Articulated Objects in Videos -> MST
% 
\subsection{Animatable 3D object generation}
Our method produces 3D keypoints along with their dense edge graphs, both of which are aligned with the coordinate system of the diffusion model and the generated multi-view images. %3D objects.
We introduce a pipeline to obtain animatable 3D objects, using the diffusion model and our 3D keypoints.

We first reconstruct 3D objects from the generated images using Gaussian Frosting \cite{guedon2024frosting}, a method built upon an efficient 3D Gaussian Splatting \cite{kerbl3dgs} framework, which outputs both 3D Gaussians and a corresponding mesh.
Next, we extract a sparse tree structure with $N-1$ edges from the 3D keypoints and their dense edge graphs by computing a Minimum Spanning Tree (MST).
The MST is constructed using a combination of predicted edge weights and Euclidean distances between keypoints.
Finally, a skinning weight $\mathbf{W}_{i,l}$ between the $i$-th vertex $\mathbf{v}_i$ and the $l$-th edge $E_l$ is computed using a Gaussian-based distance function $G$:
\begin{equation}
    \mathbf{W}_{i,l} = \frac{ G(\mathbf{v}_i, E_l) }
    {\Sigma_{j=1}^{N-1}{ G(\mathbf{v}_i, E_j) }},
\end{equation}
\begin{equation}
    G(\mathbf{v}_i, E_l) = \text{exp}(-\frac{\alpha \cdot d(v_i, E_l)^2}{2\sigma^2 }),
\end{equation}
where $d(\cdot,\cdot)$ is the minimum distance between a point and an edge, and $\sigma$, $\alpha$ are adjustable parameters.
These skinning weights and edges offer controllability in a way similar to commonly used skeletons.

\section{Experiments}
We evaluate the effectiveness of our method through extensive experiments.
These include tests on diverse types of objects across controlled, in-the-wild, and out-of-domain scenarios.
Additionally, we present animatable 3D models that are generated by our pipeline.
% Please refer to the supplements for additional quantitative and qualitative results on more diverse object categories.

\begin{table*}
\centering
\footnotesize
\begin{adjustbox}{valign=t}
\begin{minipage}{0.62\textwidth}
\caption{Quantitative comparison of 3D keypoint on the Human3.6M dataset.
* denotes results on a simplified subset with six actions.}
\label{table:h36m}
\centering
\setlength{\tabcolsep}{4pt}
\begin{tabular}{c|c|ccc|ccc}
\toprule
                             & Method                         & \#Views & \#KP & Regression & MPJPE           & N-MPJPE         & P-MPJPE        \\ \midrule
\multirow{4}{*}{\makecell{Human\\Pose}}  & Sosa \etal \cite{sosa2023self}                   & 1        & 18       & -            & -                 & -                  & 96.4           \\
                             & Kundu \etal \cite{kundu2020pgnis}                  & 1        & 18        & -            & 99.2            & -                & -                \\
                             & Kundu \etal \cite{kundu2020kinematic}                  & 1        & 18        & -            & -                 & -                 & 89.4           \\
                             & Yang \etal * \cite{yang2024mask}                 & 4        & 18        & -            & 85.6            & 85.6            & 79.3           \\ \midrule
\multirow{4}{*}{\makecell{Multi\\View}}  & \multirow{2}{*}{BKind-3D \cite{sun2023bkind3d}}      & 4        & 15     & Linear     & 125             & -               & 105            \\
                             &                                & 2        & 15     & Linear     & 155             & -               & 117            \\ \cline{2-8} 
                             & \multirow{2}{*}{Honari \etal \cite{honari2024unsupervised}} & 4        & 32     & Linear     & 120.9           & 117.9           & 93.5           \\
                             &                                & 4        & 32     & 2 hid MLP  & 73.8            & 72.6            & 63.0           \\ \midrule
\multirow{8}{*}{\makecell{Single\\View}} & Keypoint-net \cite{suwajanakorn2018keypointnet}              & 1        & 32     & 2 hid MLP  & 158.7           & 156.8           & 112.9          \\
                             & Honari \etal \cite{honari2024unsupervised}                 & 1        & 32     & 2 hid MLP  & 125.73          & 121.04          & 89.05          \\ \cline{2-8} 
                             & \multirow{4}{*}{Ours}          & 1        & 18     & Linear     & 130.58          & 127.69          & 96.83          \\
                             &                                & 1        & 18     & 2 hid MLP  & \textbf{121.34} & \textbf{118.29} & \textbf{85.26} \\
                             &                                & 1        & 32     & Linear     & 127.41          & 124.93          & 96.18          \\
                             &                                & 1        & 32     & 2 hid MLP  & \textbf{119.07} & \textbf{116.02} & \textbf{85.37} \\ \cline{2-8} 
                             & \multirow{2}{*}{Ours *}        & 1        & 18     & Linear     & 102.39          & 100.60          & 80.16          \\
                             &                                & 1        & 18     & 2 hid MLP  & \textbf{85.47}  & \textbf{84.38}  & \textbf{66.73} \\ \bottomrule
\end{tabular}
\end{minipage}
\end{adjustbox}
\hfill
\begin{adjustbox}{valign=t}
\begin{minipage}{0.355\textwidth}
\vspace{0pt}
\caption{Quantitative comparison of unsupervised 2D keypoints on CUB-200-2011 \cite{wah2011cub} dataset. We report the $L_2$ distances (normalized by image size) between GT and the regressed keypoints. Values in parentheses indicate reproduced results with the official source codes, while other results are reported as in the original papers.
% Results with $\dagger$ are reproduced without using farthest point sampling.
}
\label{table:cub}
\centering
\setlength{\tabcolsep}{4pt}
\begin{tabular}{lcc}
\toprule
Method & \begin{tabular}[c]{@{}c@{}}CUB-aligned\\ (KP=10)\end{tabular} & \begin{tabular}[c]{@{}c@{}}CUB-all\\ (KP=4)\end{tabular} \\ \midrule
SCOPS \cite{hung2019scops}                          & -                & 12.6         \\
Choudhury \etal \cite{choudhury2021unsupervised}    & -                & 9.2          \\ %\hline
Lorenz \etal \cite{lorenz2019part_disentangling}    & 3.91 (10.34)           & -            \\
ULD \cite{Thewlis_2017_ICCV, zhang2018unsupervised} & -                & 30.1         \\
Zhang \etal \cite{zhang2018unsupervised}            & 5.36             & 22.4         \\
LatentKeypointGAN \cite{he2021latentkeypointgan}    & 5.21             & 14.7         \\
GANSeg \cite{he2022ganseg}                          & 3.23 (15.73)  & 12.1         \\
Autolink \cite{he2022autolink}                      & 4.15 (8.10)        & 11.6    \\
StableKeypoints \cite{hedlin2024stable_keypoints}   & 5.06 (6.38)          & \textbf{5.4} \\
% StableKeypoints $\dagger$ \cite{hedlin2024stable_keypoints}  & 7.00    & 6.2          \\
Ours                                                & \textbf{5.16}             & 7.7          \\ \bottomrule
\end{tabular}
\end{minipage}
\end{adjustbox}
\vspace{-5pt}
\end{table*}
% \begin{figure}[t]
% \centering
% \includegraphics[trim={0 5 0 0 pt}, width=1\linewidth]{Main/Assets/results_hm36.pdf}
%    \caption{Qualitative results on the Human3.6M dataset. }
%    \label{fig:results_hm36}
%    \vspace{-5pt}
% \end{figure}
\begin{figure*}[t]
\centering
\includegraphics[trim={0 20 0 0 pt}, width=1\linewidth]{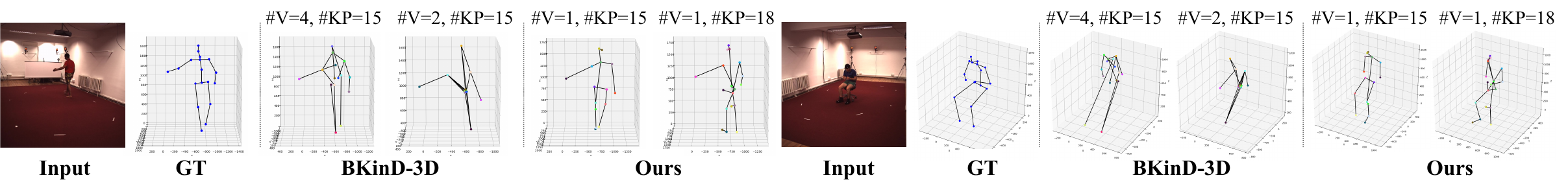}
   \caption{Qualitative comparison on the Human3.6M dataset. }
   \label{fig:results_hm36}
   \vspace{-10pt}
\end{figure*}

%% [paragraph or subsection?]
\textblock{Implementation details.}
We use pretrained SV3D-p~\cite{voleti2024sv3d} as our multi-view diffusion backbone, which generates 21 geometrically consistent views conditioned on the input image and camera parameters. 
Among these, we use $K = 4$ views during training: the input view and 3 additional generated views.
Intermediate decoder features are extracted at diffusion timestep $\tau = 500$ from all layers of the U-Net, with a total of $T=1000$ denoising steps.
As preprocessing, we extract the foreground object mask using the Grounded SAM \cite{ren2024grounded} before passing the image to the diffusion model.
We train all modules except the reconstruction network using the AdamW optimizer with a learning rate of $1 \times 10^{-4}$, and the reconstruction network with a learning rate of $1 \times 10^{-3}$.
We train our model for 20{,}000 steps on two NVIDIA A100 GPUs (40GB), with a total batch size of 4.
We set the grid resolution to $M = 72$ and use $N = 18$ keypoints by default.
For the multi-view reconstruction loss, we set $\lambda_{\text{vgg}} = 1.0$ and $\lambda_{\text{mask}} = 0.5$.

\subsection{Datasets}
We train our model on diverse object categories, spanning both humans and animals.

\textblock{Human3.6M}~\cite{ionescu2013h36m} is a large-scale human activity dataset with multi-view videos recorded from four calibrated cameras. 
Following standard protocols \cite{iskakov2019learnable, kocabas2019self}, we use subjects 1, 5, 6, 7, and 8 for training, and 9 and 11 for testing. 
While the dataset provides multi-view videos, we treat each frame as an independent single-view image, without using camera parameters or temporal information during training.

\textblock{CUB-200-2011} \cite{wah2011cub} contains 11,788 images across 200 bird species.
Following prior works, we evaluate on two standard protocols: \textit{CUB-aligned} \cite{lorenz2019part_disentangling} and \textit{CUB-all} \cite{choudhury2021unsupervised}.
The CUB-aligned version consists of pre-aligned images in which all birds face left, and seabird categories are excluded.
In contrast, CUB-all uses unaligned images cropped with the provided bounding boxes, and we follow the train/val/test split of~\cite{choudhury2021unsupervised}.

\textblock{Stanford Dogs}~\cite{stanford_dogs} consists of over 22{,}000 images of 120 dog breeds. We split the dataset by breeds, using 92 breeds for training and 28 unseen breeds for testing. 
We subsample the dataset by selecting images in which the full body is visible, resulting in 3{,}000 training images.

\textblock{DAVIS} \cite{Perazzi2016davis}, \textbf{GSO} \cite{downs2022google}, and \textbf{AP-10K} \cite{yu2021ap} are used to evaluate the in-the-wild and out-of-domain generalization of our model.
These datasets are not used during training; they are employed for qualitative evaluation only.

\subsection{3D keypoint estimation}
We assess our unsupervised 3D keypoint accuracy using the Human3.6M dataset with ground-truth 3D annotations.

\textblock{Regression.}
To evaluate the accuracy of our unsupervised 3D keypoints, we follow the standard protocol adopted in prior unsupervised keypoint discovery works~\cite{jakab2018unsupervised, sun2023bkind3d,honari2024unsupervised}. 
Specifically, we train a regressor that maps the predicted keypoints to ground-truth 3D annotations using the training split of data.
We experiment with two types of regressors: (1) a linear regression model without a bias term, and (2) a two hidden layer MLP with hidden dimensions of (50, 50) and ReLU activations.
% Importantly, only the regressor is supervised using ground-truth annotations, while the keypoints themselves are learned in a fully unsupervised manner.

\begin{figure*}
\centering
\includegraphics[trim={0 20 0 25pt}, width=1\linewidth]{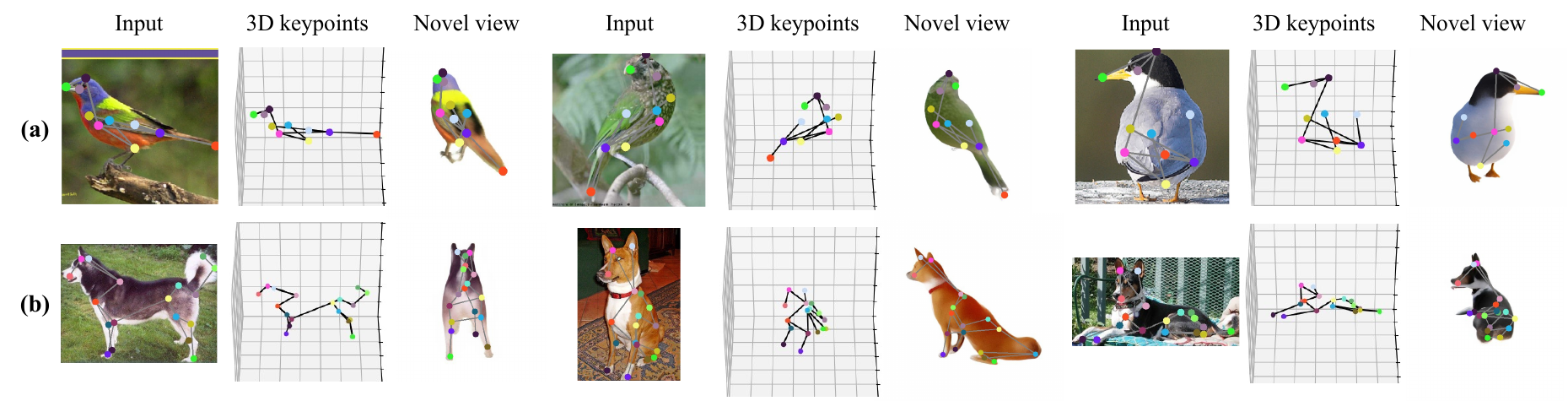}
\vspace{-5pt}
\caption{Qualitative results on the (a) CUB-200-2011 and (b) Stanford Dogs datasets. }
\label{fig:results_animal}
\vspace{-5pt}
\end{figure*}

\begin{figure*}
\centering
\includegraphics[trim={0 5 0 0pt}, width=0.98\linewidth]{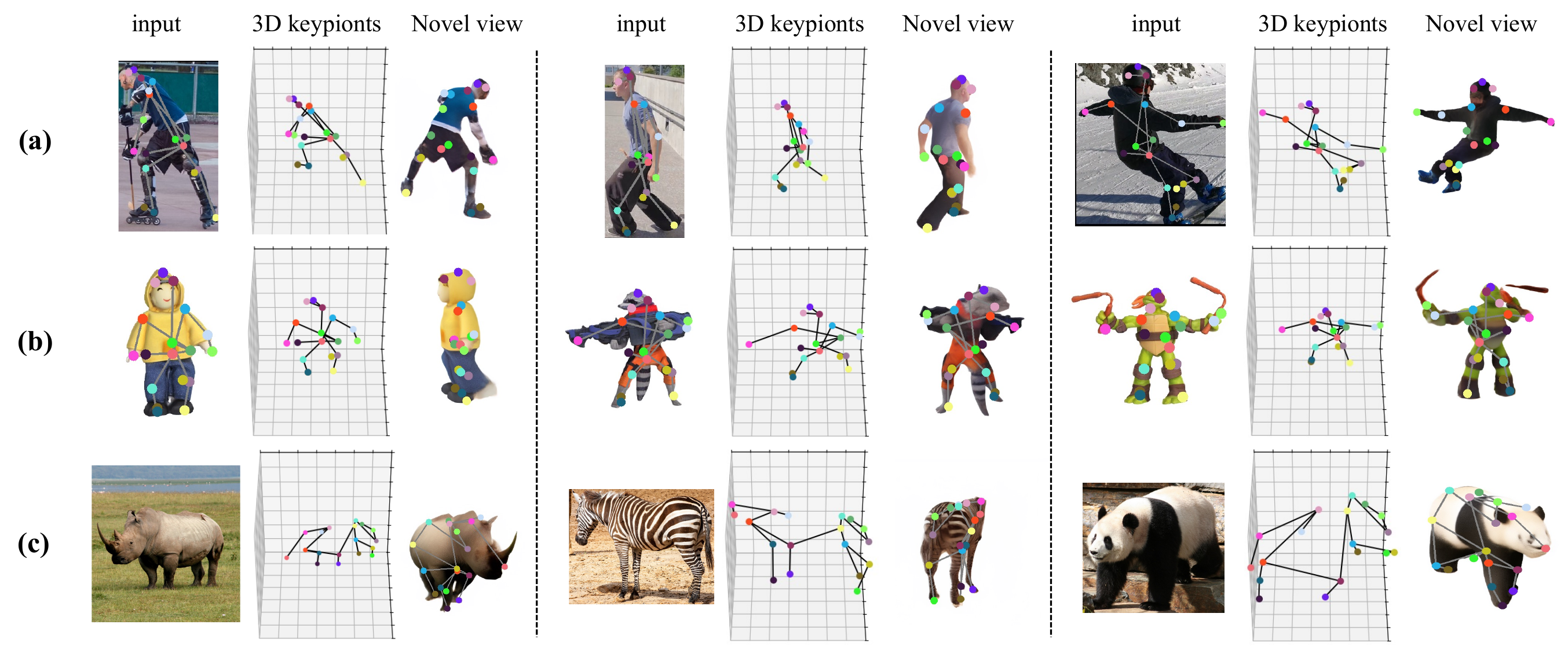}
\vspace{-5pt}
\caption{Out-of-domain generalization results. (a) In-the-wild DAVIS results and (b) out-of-domain GSO results using a model trained on Human3.6M.
(c) AP-10K results using a model trained on Stanford Dogs.}
\label{fig:results_wild}
\vspace{-10pt}
\end{figure*}

\textblock{Metrics.}
We evaluate 3D human pose estimation performance using standard metrics:
Mean Per Joint Position Error (\textbf{MPJPE}) measures the average L2 distance between predicted and ground-truth 3D keypoints, computed in millimeters.
Normalized MPJPE (\textbf{N-MPJPE}) scales the predicted pose to match the scale of the ground-truth pose before computing the MPJPE. 
Procrustes-aligned MPJPE (\textbf{P-MPJPE}) further aligns the predicted pose to the ground-truth via a Procrustes transformation (including translation, rotation, and uniform scaling) and then computes MPJPE.
%, which removes global translation, rotation, and scale. 

\textblock{Baselines.}
We compare our method against prior works on unsupervised 3D keypoint estimation. 
As the unsupervised monocular setup is relatively underexplored, we include the following relevant baselines for comparison:
% We compare our method with prior works on unsupervised 3D keypoint estimation. 
% While our focus is on discovering 3D keypoints from monocular input without any multi-view or 3D supervision, this setting remains relatively underexplored. For thorough comparison, we consider the following relevant baselines:
(1) \textit{unsupervised monocular human pose estimation methods}, which do not rely on paired 3D annotations. These methods instead leverage weak supervision such as unpaired 2D poses~\cite{sosa2023self}, unpaired 3D poses~\cite{kundu2020pgnis}, kinematic constraints like joint connectivity and bone-length ratios~\cite{kundu2020kinematic,yang2024mask};
(2) \textit{unsupervised multi-view 3D keypoint estimation methods}, which require synchronized multi-view images and known camera parameters for both training and inference~\cite{sun2023bkind3d, honari2024unsupervised};
(3) \textit{monocular inference with multi-view training}, where training uses multi-view data, but inference is performed on a single image \cite{suwajanakorn2018keypointnet, honari2024unsupervised}.

\textblock{Quantitative results.}
\Tref{table:h36m} summarizes the quantitative performance of our method compared to existing baselines.
Our method outperforms all unsupervised single-view baselines.
It even achieves competitive results with multi-view baselines with only using a single-view image, surpassing the performance of BKinD-3D~\cite{sun2023bkind3d} with 2 views.
We emphasize that the single-view setup is significantly more challenging than the multi-view setup, due to the large portions of occluded regions and depth ambiguity.
Notably, our method also achieves improved P-MPJPE scores compared to monocular human pose estimation methods employing human-specific priors.
These results demonstrate the effectiveness of our approach in discovering accurate and 3D keypoints from monocular images without relying on multi-view supervision or human-specific priors.

\textblock{Qualitative results.}
We present a qualitative comparison on the Human3.6M dataset in \fref{fig:results_hm36}.
Note that the visualized keypoints are the raw model predictions, not the regressed ones.
Compared with BKinD-3D, our method yields more fine-grained and accurate keypoints, especially for difficult joints such as the knees, which are often missed by BKinD-3D. This gap becomes more evident as the number of input views decreases. In particular, whereas BKinD-3D shows a significant performance drop with only two input views, our method remains robust with a single view.

\subsection{Animal keypoint estimation}
We train our model on diverse animal categories, including birds and dogs, using the CUB-200-2011 \cite{wah2011cub} and Stanford Dogs \cite{stanford_dogs} datasets.
Both datasets consist of single images captured in natural environments (not multi-view or video) and do not provide 3D annotations.

\textblock{Qualitative results.}
\fref{fig:results_animal} shows qualitative results on CUB-200-2011 and Stanford Dogs.
Across bird species, our method reliably captures semantic parts such as the beak (green) and tail (orange), along with 3D structure under varied poses.
For dogs, our method produces consistent keypoints even under challenging articulations (\eg sitting) and in the presence of occlusions.

\textblock{Quantitative results.}
We evaluate our method on the CUB-200-2011 dataset and compare it with prior unsupervised 2D keypoint methods.
Due to the absence of 3D annotations, we instead project our predicted 3D keypoints to 2D, and follow the 2D evaluation protocols on both CUB-aligned and CUB-all benchmarks.
As shown in \Tref{table:cub}, although these evaluations do not fully reflect the strengths of our 3D predictions, our projected keypoints still match the performance comparable to 2D baselines.
We also note that most of the CUB-aligned scores of 2D baselines reported in their original papers cannot be reproduced, even when using their official source codes.
Please refer to the supplements for more details and comparisons with 2D baselines.

% As shown in \Tref{table:cub}, our projected keypoints achieve performance comparable to 2D baselines.
% It is worth noting that these 2D evaluations do not fully assess the underlying 3D structures that our method uniquely provides.
% Please refer to the supplements for more details and comparisons with 2D baselines.
% Following the standard protocol, we project our predicted 3D keypoints to 2D, regress them to ground-truth 2D keypoints using linear regression, and compute the normalized $L_2$ distance weighted by the ground-truth visibility signal.
% Our method achieves competitive performance on both the CUB-aligned and CUB-all benchmarks. %, despite not being explicitly designed for 2D localization.
% It is worth noting that the metric only evaluates visible keypoints and does not account for occluded regions.
% In contrast, our model demonstrates strong occlusion robustness as shown in the qualitative results.

% % Since our model predicts 3D keypoints, which cannot be directly compared to 2D baselines, we project them onto the image plane and regress the projected points to the ground-truth 2D keypoints following the standard protocol.
% We report the normalized $L_2$ distance (scaled by image size) multiplied by the visibility of each keypoint. The results are presented in Table~\ref{table:cub}.

\subsection{Generalization}
We further evaluate our model under several out-of-domain scenarios to test its robustness beyond the training distribution, using models trained on the Human3.6M and Stanford Dogs datasets.
\textit{(1) In-the-wild scenario.}
To assess generalization in real-world conditions, we present qualitative results on the DAVIS dataset in \fref{fig:results_wild} (a).
Although the model is trained only on the indoor Human3.6M dataset with five subjects, it generalizes well to in-the-wild inputs.
\textit{(2) Out-of-domain scenario.}
We further evaluate the same model on animated characters in GSO dataset.
As shown in \fref{fig:results_wild} (b), the model produces consistent keypoints across these previously unseen categories.
% \textit{(3) Cross-species scenario.}
We also evaluate a model trained on Stanford Dogs by applying it to different animal species—including rhinoceroses, zebras, and pandas (\fref{fig:results_wild} (c)).
Despite the large variations in shape, appearance, and limb structure, our method consistently predicts semantically meaningful keypoints, demonstrating its robustness across species.

% First, we test our model trained on Human3.6M dataset on the in-the-wild human videos in DAVIS dataset and animation characters in GSO dataset.

% \textblock{Generalization to animal domains.}
% We train our model on the Stanford Dogs dataset, a single-image in-the-wild dataset containing various dog breeds in natural environments. 
% \fref{fig:results_animal} (left) shows qualitative results on unseen dog breeds, where our method produces consistent keypoints even under challenging poses (\eg, sitting) and viewpoints (\eg, back view).
% To further assess cross-category generalization, we evaluate our model on out-of-domain animals including rhinoceroses and zebras (\fref{fig:results_animal} (right)), as well as giraffes (\fref{fig:teaser}). 
% Despite the large variations in shape, appearance, and a limb structure, our method consistently predicts semantically meaningful keypoints, demonstrating its robustness across species without requiring animal-specific supervision.

%%% [in] 시간 되면 여기 한번 다시 보자..
\begin{figure*}[t]
\centering
\includegraphics[trim={0 25 0 25pt}, width=1\linewidth]{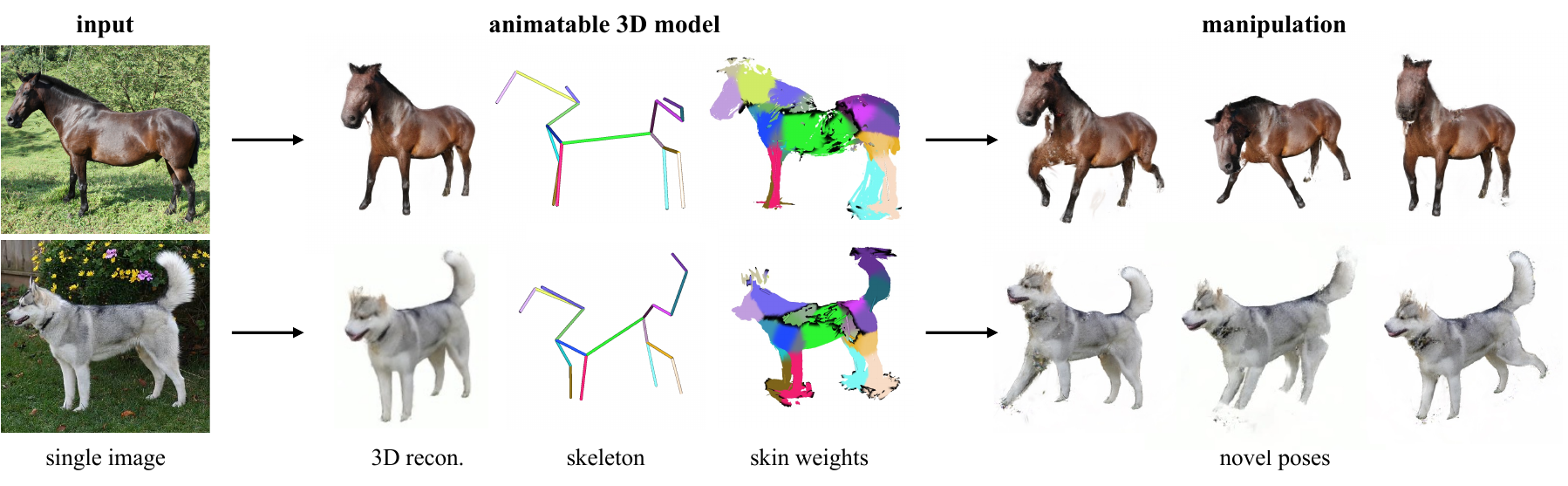}
\vspace{-5pt}
\caption{Animatable 3D model results.}
%% for dummy
\label{fig:results_animation}
\vspace{-10pt}
\end{figure*}
\subsection{Manipulating generated 3D objects}
Our pipeline enables manipulation of generated 3D objects by leveraging predicted 3D keypoints and edge graphs.
% , which are inherently aligned with the diffusion model's coordinate system.
In parallel with keypoint estimations, the same diffusion model generates multi-view images, which are used to reconstruct a mesh along with 3D Gaussian Splatting~\cite{kerbl3dgs, guedon2024frosting}.
% In parallel with keypoint estimation, the same diffusion model generates multi-view images, which are used to reconstruct a mesh along with Gaussian Splatting~\cite{kerbl3dgs, guedon2024frosting}.
Since our predictions are aligned with the reconstructed objects, we can directly manipulate these objects using them without additional registration.
\fref{fig:results_animation} shows examples of reconstructed objects, predicted skeletons, computed skinning weights, and manipulation results.
These results showcase that our KeyDiff3D allows articulation and deformation of the generated 3D objects, without requiring object-specific skeleton design or manual rigging.

\subsection{Ablation study}
\begin{table}
\caption{Ablation results on (a) 2D feature backbones, (b) 3D lifting strategies, and (c) the number of virtual viewpoints. All results are reported using 2-layer MLP regression.}
\resizebox{1\columnwidth}{!}{
\setlength{\tabcolsep}{10pt}
\begin{tabular}{lccc}
\toprule
\multicolumn{1}{c}{}    & MPJPE           & N-MPJPE         & P-MPJPE        \\ \midrule
\multicolumn{4}{c}{(a) Backbone}                                             \\ \midrule
ResNet50 \cite{he2016resnet}               & 138.97          & 136.34          & 101.55         \\
CLIP \cite{radford2021clip}                   & 143.23          & 139.96          & 103.13         \\
DINOv2 \cite{oquab2023dinov2}                 & 136.17          & 133.53          & 101.91         \\
Ours (SV3D)             & \textbf{121.34} & \textbf{118.29} & \textbf{85.26} \\ \midrule
\multicolumn{4}{c}{(b) 3D lifting}                                       \\ \midrule
2D kpts  $\rightarrow$ triangulation & 129.63          & 126.96          & 93.68          \\
3D features $\rightarrow$ 3D kpts                 & \textbf{121.34} & \textbf{118.29} & \textbf{85.26} \\ \midrule
\multicolumn{4}{c}{(c) Number of virtual viewpoints}                         \\ \midrule
1 viewpoint (input)          & 166.29          & 157.94          & 104.22         \\
2 viewpoints                      & 132.36          & 129.41          & 92.36          \\
3 viewpoints                      & 121.58          & 118.75          & 86.59          \\
4 viewpoints                      & 121.34          & 118.29          & 85.26          \\
5 viewpoints                      & \textbf{119.77}          & \textbf{117.42}          & \textbf{82.94} \\ \bottomrule
\end{tabular}
}
\label{table:ablation_backbone}
\vspace{-10pt}
\end{table}

To further probe the effectiveness of our modules, we conduct ablation studies on Human 3.6M.
Please refer to the supplements for additional ablation results.

\textblock{Diffusion model as a 3D feature extractor.}
We first evaluate the effectiveness of employing a multi-view diffusion model as a foundational 3D feature extractor.
For ablation, we replace our diffusion-based features with those extracted from generated multi-view images using ResNet50 \cite{he2016resnet}, CLIP \cite{radford2021clip} image encoder, and DINOv2 \cite{oquab2023dinov2}.
All extracted 2D features are unprojected and aggregated into a 3D volume following the same pipeline as our main method.
As shown in \Tref{table:ablation_backbone} (a), multi-view diffusion features consistently outperform conventional 2D foundational backbones.
This suggests that (1) the multi-view diffusion model encodes stronger 3D geometric awareness than prior 2D-based feature extractors, and (2) its intermediate multi-view features contain richer geometric priors than the rendered multi-view images themselves.

% % %%% [in] 여기는 분량 부족하면 다시 바꾸자
% Simply using diffusion-generated images as multi-view inputs does not yield accurate keypoints.
% These results highlight the importance of multi-view feature extractors, indicating that diffusion models can provide multi-view features with strong geometry priors.
% Our model effectively transforms these geometric priors into 3D features, resulting in a significant improvement in accuracy.
% These results indicate that multi-view diffusion models provide multi-view features encoded with strong geometric priors, which leads to the significant improvement in accuracy.
% Qualitatively, diffusion-based keypoints are more spatially coherent and better aligned with human-understandable structures.

\textblock{3D lifting strategies.} To validate the effectiveness of our 3D keypoint extraction network, which unprojects 2D multi-view features into a 3D volumetric space, we compare it with a variant that employs a 2D keypoint detector followed by triangulation.
Specifically, we train a 2D keypoint detector using the same multi-view diffusion features as our method, and estimate 3D keypoints by triangulating the predicted 2D keypoints across diffusion-generated views using the corresponding camera parameters.
As shown in \Tref{table:ablation_backbone} (b), this replacement leads to a noticeable drop in accuracy.
Although diffusion-generated multi-view images can be combined with existing 2D-based keypoint estimation methods, our results demonstrate that explicitly modeling keypoints in 3D space with a volumetric network yields more consistent and geometrically accurate predictions.

\textblock{Number of virtual viewpoints.}
Finally, we analyze the effect of the number of diffusion-generated views. % used during training and inference 
As shown in \Tref{table:ablation_backbone} (c), using only the input image ($K{=}1$) results in substantially lower accuracy due to insufficient geometric cues for 3D localization.
Introducing even one generated view yields large improvements, with performance gradually saturating beyond $K{=}3$.
We set $K{=}4$ in all experiments to balance accuracy and efficiency.

\section{Conclusion}

We presented \textbf{KeyDiff3D}, a framework for predicting 3D keypoints from a single image without requiring 3D or multi-view supervision. 
By leveraging pretrained multi-view diffusion models as geometry-aware feature extractors in both training and feature extraction, our method exhibits strong performance across a wide range of scenarios.
% including in-the-wild and out-of-domain generalization to unseen categories such as animals and non-human objects.
% By leveraging pretrained multi-view diffusion models as geometry-aware feature extractors, our method constructs 3D volumes from generated views and learns keypoint representations via self-supervised reconstruction. 
% Experiments demonstrate strong performance across a range of scenarios, including in-the-wild and out-of-domain generalization to unseen categories such as animals and non-human objects. 
These results highlight the potential of diffusion-based priors for self-supervised 3D understanding, opening up new directions for scalable and generalizable geometry learning.

\clearpage
\section*{Acknowledgments}
This work was supported by Institute of Information \& communications Technology Planning \& Evaluation (IITP) grant funded by the Korea government(MSIT) (No.RS-2022-II220124, No. RS-2024-00457882), and Artificial Intelligence Graduate School Program grant funded by Yonsei University (RS-2020-11201361).
{
    \small
    \bibliographystyle{ieeenat_fullname}
    \bibliography{main}
}


\newpage
\appendix

\section{CUB-200-2011 comparison}
\paragraph{Baseline reproduction.}
Most unsupervised 2D keypoint methods \cite{lorenz2019part_disentangling,he2022ganseg,he2022autolink,hedlin2024stable_keypoints} report their performance on the CUB-align protocol introduced in \cite{lorenz2019part_disentangling}.
However, we note that it is challenging to ensure a fair and fully controlled comparison under this setup, as the preprocessing code, train/test splits, and the preprocessed dataset have not been publicly released by these methods.
% However, we note that it is challenging to ensure a fair and fully controlled comparison under this setup, as neither the preprocessing code nor the preprocessed CUB-align dataset has been publicly released by these methods.
The original paper \cite{lorenz2019part_disentangling} specifies that images are roughly cropped using the provided landmarks, which can potentially lead to multiple plausible variants of the preprocessed dataset.
Since normalized keypoint distances are highly sensitive to the crop ratio used during preprocessing, such inconsistencies make it challenging to directly compare our results with previously reported CUB-align scores.

To ensure fair comparisons under the identical evaluation protocol, we re-train baseline models that report CUB-align results in their main papers and provide official implementations \cite{lorenz2019part_disentangling, he2022ganseg, he2022autolink, hedlin2024stable_keypoints}.
For consistency, we use the publicly available preprocessed data from \cite{braun2020part_reproduce} and train these baselines using the same preprocessing and train/test split.
For each baseline, we report both the originally published numbers and the results reproduced using their official code.

\paragraph{Reproduced results and quantitative comparisons.}
\Tref{table:supp_cub} (left) reports the reproduced results on the CUB-align dataset.
Under the same evaluation protocol, our method outperforms other 2D baselines in terms of 2D accuracy, demonstrating that our results not only capture 3D consistent structures but also yield accurate 2D projected keypoints.
% Our method shows a slight improvement when compared to its CUB-all score, likely because the increased number of keypoints allows for more fine-grained localization.
% Overall, the reproduced CUB-align scores are close to their reported scores under the CUB-all setup.
% Overall, the reproduced results align with those reported under the CUB-all setting, with modest improvements across most baselines.
% We attribute these gains primarily to the larger keypoint set used during training, which provides richer supervisory signals for fine-grained localization.
The reproduced scores of StableKeypoints closely match their CUB-all (KP=4) results, likely due to their fixed-token training strategy: the model always learns with 25 top-k tokens chosen from 500 learnable keypoint embeddings, and selects the final subset via furthest point sampling, leading to consistent performance across different keypoint counts.
% Under this unified evaluation setting, our method achieves the highest accuracy among the baselines.

We additionally evaluate on the CUB-all dataset with 10 keypoints (KP=10), since 4 keypoints are often insufficient to capture overall structures.
As shown in \Tref{table:supp_cub} (right), our method achieves further improvements with more keypoints and outperforms previous 2D baselines. Similar to CUB-align results, increasing the number of keypoints of StableKeypoints does not lead to the meaningful improvement due to its fixed-token sampling strategy.
These results indicate that our method accurately captures structures of the objects in both 3D and 2D projections.

\begin{table}
\caption{Reproduced results on CUB-align and CUB-all dataset. For fair comparison, reproduced results are obtained using the identical preprocessed dataset with same train/test splits, and official source codes provided by the authors. We report reproduced results for \textit{CUB-all}.}
\centering
\footnotesize
\setlength{\tabcolsep}{5pt}
\begin{tabular}{l|cc|cc}
\toprule
 & \multicolumn{2}{c|}{\textit{CUB-align} (KP=10)} & \multicolumn{2}{c}{\textit{CUB-all}} \\ %\cline{2-3}
 % & \multicolumn{2}{c|}{KP=10} & KP=4 & KP=10 \\
 % & Original & Reproduced & \multicolumn{2}{c}{Reproduced} \\
 & Original & Reproduced & KP=4 & KP=10 \\
 \midrule
Lorenz \etal \cite{lorenz2019part_disentangling} & 3.91 & 10.34 & - & - \\
GANSeg \cite{he2022ganseg} & \textbf{3.23} & 15.73 & - & - \\
Autolink \cite{he2022autolink} & 4.15 & 8.10 & 13.0 & 11.7 \\
StableKeypoints \cite{hedlin2024stable_keypoints} & 5.06 & 6.38 & \textbf{5.6} & 6.0 \\
Ours & - & \textbf{5.16} & 7.7 & \textbf{5.5} \\
\bottomrule
\end{tabular}
\label{table:supp_cub}
\vspace{-5pt}
\end{table}

\begin{figure}
\begin{center}
\includegraphics[width=1\linewidth]{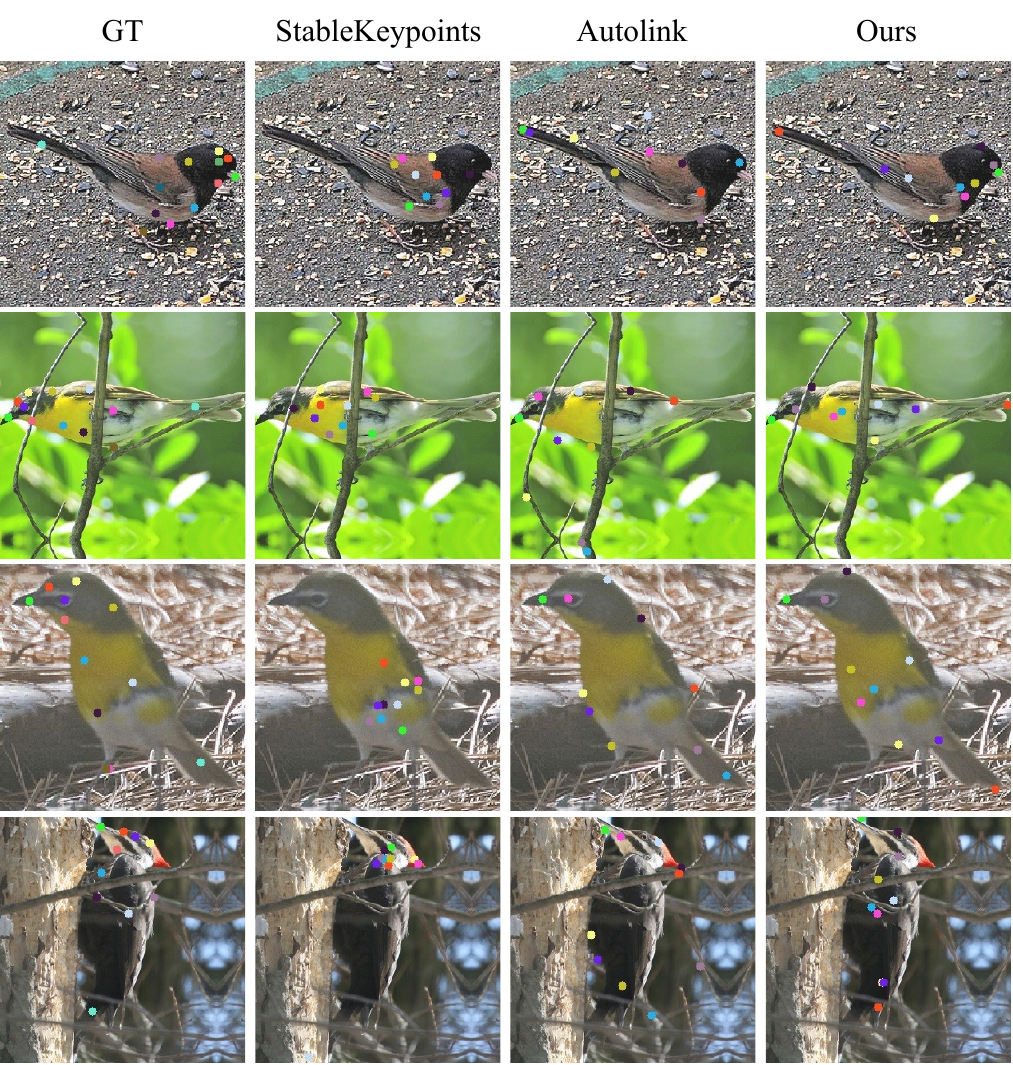}
\end{center}
   \caption{Qualitative comparison with unsupervised 2D keypoint estimation methods on CUB-200-2011 dataset (CUB-align). }
   \label{fig:supple_cub_comparison}
   \vspace{-10pt}
\end{figure}
\paragraph{Qualitative comparsions.}
We present qualitative comparisons of 2D keypoint predictions in \fref{fig:supple_cub_comparison}.
StableKeypoints tends to produce individually consistent keypoints but often lacks a coherent global structure across the object.
AutoLink, while effective in capturing overall structure, struggles to maintain consistency under pose variations; for instance, in the first two rows, it frequently confuses the head and tail regions, assigning different keypoints to the tail depending on the viewpoint.
In contrast, our method produces stable and structurally coherent keypoints across diverse poses and viewing directions.
We attribute this robustness to the use of multi-view diffusion features within a 3D representation, which enables more geometry-aware keypoint reasoning.

\begin{figure}[t]
\begin{center}
\includegraphics[width=1\linewidth]{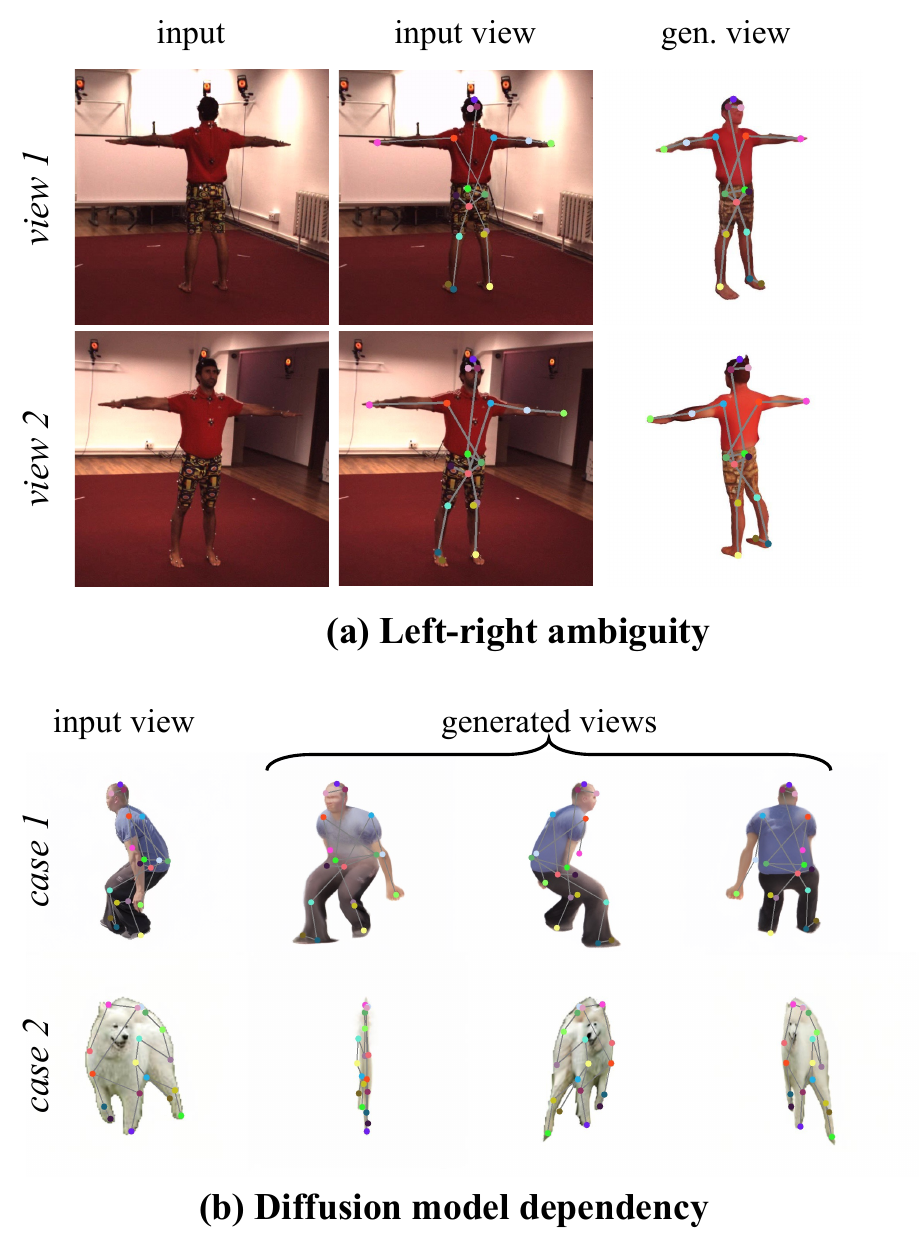}
\end{center}
   \vspace{-20pt}
   \caption{Failure cases. }
   \label{fig:supple_failures}
   \vspace{-10pt}
\end{figure}
\begin{table}[]
\caption{Additional ablation results on the Human3.6M dataset. `default' indicates the default configuration used in our main experiments.}
\centering
\resizebox{1\columnwidth}{!}{
\setlength{\tabcolsep}{8pt}

\begin{tabular}{lccc}
\toprule
\multicolumn{1}{c}{} & MPJPE                      & N-MPJPE                    & P-MPJPE                   \\ \midrule
\multicolumn{4}{c}{(a) Number of keypoints}                                                                \\ \midrule
$N=16$                   & 127.99                     & 123.93                     & 85.57                     \\
$N=18$ (default)         & 121.34                     & 118.29                     & 85.26                     \\
$N=32$                   & 119.07                     & 116.02                     & 85.37                     \\
$N=48$                   & \textbf{116.45}            & \textbf{113.29}            & \textbf{81.30}            \\ \midrule
\multicolumn{4}{c}{(b) Diffusion timestep}                         \\ \midrule
$\tau=300$           & 126.97          & 122.79          & 88.54          \\
$\tau=500$ (default) & \textbf{121.34} & \textbf{118.29} & \textbf{85.26} \\
$\tau=700$           & 130.16          & 125.68          & 88.32          \\
$\tau=900$           & 131.26          & 127.13          & 88.62          \\ \midrule
\multicolumn{4}{c}{(c) Number of training images}                                                          \\ \midrule
2,200 images         & 134.26                     & 130.01                     & 90.71                     \\
4,400 images         & 130.85                     & 126.74                     & 88.96                     \\
6,600 images         & 127.42                     & 123.88                     & 85.50                     \\
8,800 images (default)     & \textbf{121.34}            & \textbf{118.29}            & \textbf{85.26}            \\ \midrule
\multicolumn{4}{c}{(d) Affine transformation} \\ \midrule
no affine & 134.54 & 130.69 & 95.09 \\
random affine (default) & \textbf{121.34} & \textbf{118.29} & \textbf{85.26} \\ \midrule 
\multicolumn{4}{c}{(e) View misalignment}                         \\ \midrule
camera noise $\sigma$=0.0° (default)  & 121.34 & 118.29 & \textbf{85.26} \\
camera noise $\sigma$=0.2° & \textbf{121.30} & \textbf{118.18}  & 85.29   \\
camera noise $\sigma$=0.5° & 124.39          & 121.26          & 86.64          \\
camera noise $\sigma$=1.0° & 133.88          & 130.68          & 92.85          \\  \midrule
\multicolumn{4}{c}{(f) View aggregation}                         \\ \midrule
softmax (default) & \textbf{121.34} & \textbf{118.29} & \textbf{85.26} \\
visibility & 127.12 & 124.29 & 86.30 \\ \bottomrule

\end{tabular}}\label{table:supp_ablation}
\end{table}

% \begin{table}[]
% \caption{Additional ablation results on the Human 3.6M dataset. }
% \centering

% \begin{tabular}{c|ccc}
% \toprule
%                             & MPJPE           & N-MPJPE         & P-MPJPE        \\
% \midrule
% 2D keypoints + triangulation & 129.63          & 126.96          & 93.68          \\
% 3D keypoints          & \textbf{121.34}          & \textbf{118.29}          & \textbf{85.26}  \\
% \hline
% %%5 number of keypoints
% $N=16$ & 127.99          & 123.93          & 85.57          \\
% $N=18$ (default) & 121.34          & 118.29          & 85.26          \\
% $N=32$ & 119.07          & 116.02          & 85.37          \\
% $N=48$ & \textbf{116.45}          & \textbf{113.29}           & \textbf{81.30}        \\
% \hline

% %%% diffusion timesteps
% $\tau=1$ & 129.63          & 126.96          & 93.68          \\
% $\tau=1$ & 129.63          & 126.96          & 93.68          \\
% $\tau=1$ & 129.63          & 126.96          & 93.68          \\
% \hline
% %%% training data
% 2,200 training images & 134.26          & 130.01          & 90.71          \\
% 4,400 training images & 130.85          & 126.74          & 88.96          \\
% 6,600 training images & 127.42          & 123.88          & 85.50          \\
% 8,800 training images (default) & \textbf{121.34}          & \textbf{118.29}          & \textbf{85.26}          \\
% \bottomrule
% \end{tabular}
% \label{table:supp_ablation}
% \end{table}

\section{Failure cases}
We illustrate our failure cases in \fref{fig:supple_failures}.
First, our method suffers from left–right ambiguity, a common issue in self-supervised keypoint learning where the model has no inherent cue to distinguish between the left and right sides \cite{suwajanakorn2018keypointnet, he2022autolink, yang2024mask}.
Although our network produces consistent keypoint predictions across input and diffusion-generated views, our model has difficulty resolving the left-right ambiguity present in the input image itself.
As shown in \fref{fig:supple_failures} (a), the model predicts the same hand (e.g., left hand) regardless of whether the input shows the front or back of the subject.

Second, since our model depends on features extracted from a pretrained diffusion model, its performance may degrade in regions where the diffusion model fails to generate plausible content.
For example, as shown in \textit{case 1} of \fref{fig:supple_failures} (b), the SV3D model sometimes fails to reconstruct occluded regions such as the right hand, resulting in inaccurate predictions.
In another failure mode (\textit{case 2}), SV3D produces inconsistent multi-view images—for instance, by flipping the subject—making it difficult for our network to predict volumetric 3D keypoints.
% Despite these limitations, our method remains robust in most cases and outperforms prior work under limited supervision and occluded scenarios.

\section{Additional ablation study}
We present additional ablation results for a more comprehensive analysis of our method.
Consistent with the main paper, all ablations are conducted on the Human3.6M dataset, and we report the results with a 2-hidden-layer MLP regression.

\begin{figure*}[t]
\begin{center}
\includegraphics[width=1\linewidth]{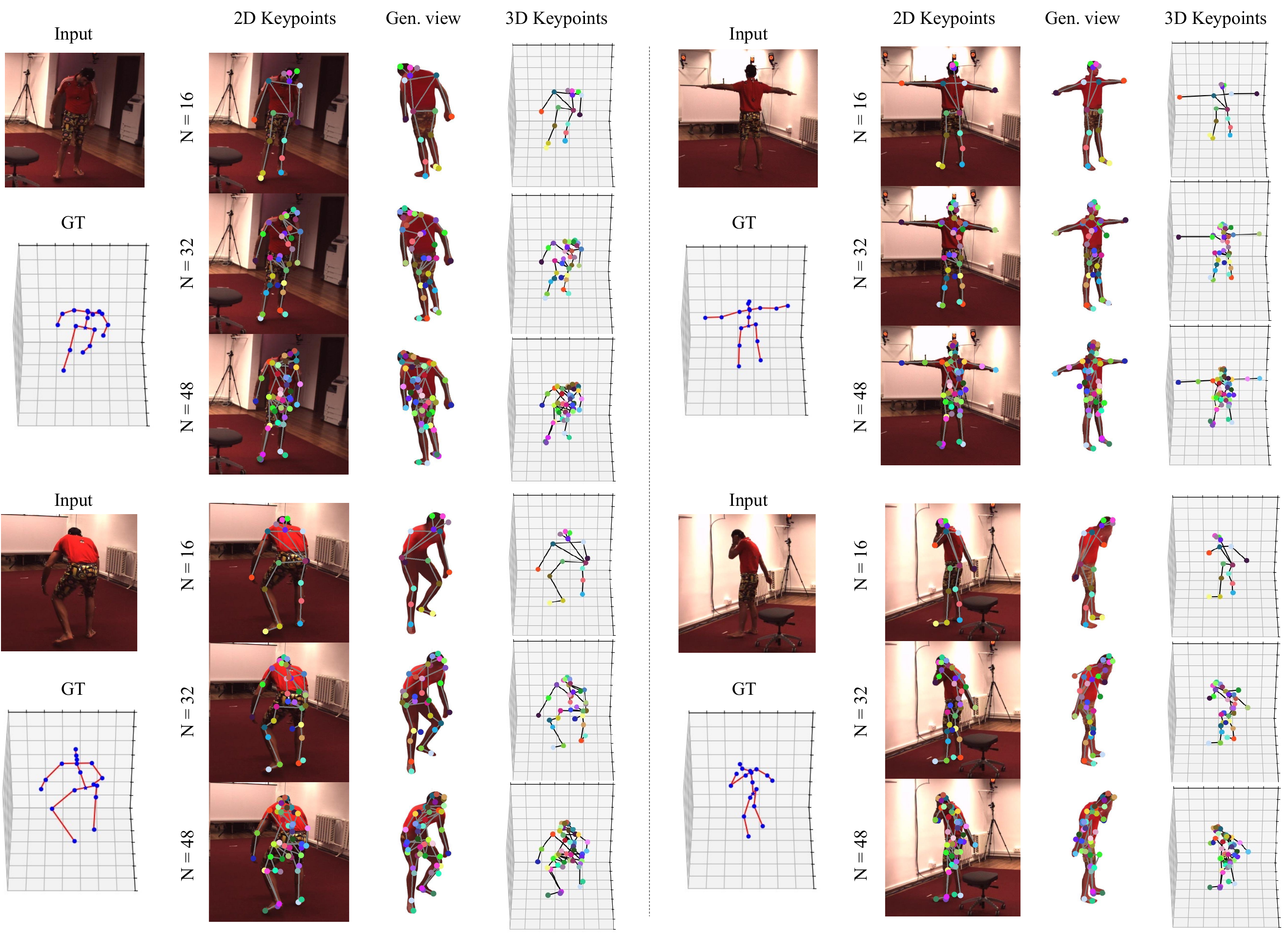}
\end{center}
   \caption{Keypoint prediction results according to the number of keypoints.}
   \label{fig:supple_keypoints}
   \vspace{0pt}
\end{figure*}

\paragraph{Number of keypoints.}
We test the performance of our model with a varying number of keypoints.
As reported in \Tref{table:supp_ablation} (a), the performance of our model improves as the number of keypoints increases.
However, as employing too many keypoints could reduce interpretability, we set the default number of keypoints $N=18$ in our main model.
This trade-off between accuracy and interpretability can also be seen in \fref{fig:supple_keypoints}.

\paragraph{Diffusion timesteps.}
We analyze the performance of our model with various diffusion timesteps $\tau$, which are used for extracting features from the diffusion U-Net backbone.
\Tref{table:supp_ablation} (b) presents the ablation results on diffusion timesteps.
We observe that the timestep $\tau = 500$ yields the best results, which we set as our default configuration.

\paragraph{Number of training data.}
We investigate the effects of the number of training images on the performance of our model.
% As shown in \Tref{table:supp_ablation} (d), our model still predicts accurate keypoints with using only 2,200 training images.
 As shown in \Tref{table:supp_ablation} (c), our model delivers strong performance with fewer than 10K training images, whereas other baselines typically require at least 30K to achieve comparable results.
Notably, our method only requires a set of unconstrained single-view images -- without any camera parameters, temporal information, or manual annotations -- which are much easier to acquire than multi-view images or 3D annotations.

\paragraph{Affine transformation.}
We further ablate the effect of applying an affine transformation to the input image before feeding it into the reconstruction network. As shown in \Tref{table:supp_ablation} (d), removing this transformation leads to a substantial performance drop. We attribute this degradation to trivial solutions in reconstruction training: without the transformation, the reconstruction network tends to overfit to the input appearance and bypass the structural cues derived from the predicted keypoints. Applying the transformation weakens this shortcut and encourages the model to rely on structure-aware signals for reconstruction, thereby improving performance.

\paragraph{View misalignment.}
We evaluate the robustness of our method to view misalignment by perturbing the camera rotations used for feature unprojection with Gaussian noise. As shown in \Tref{table:supp_ablation} (e), the performance degrades only slightly under mild perturbations, indicating that our method is robust to moderate inaccuracies in the poses of diffusion-generated views.

\paragraph{View aggregation.}
We further compare our default softmax-based view aggregation with a visibility-aware alternative based on depth maps predicted by VGGT. As shown in \Tref{table:supp_ablation} (f), the visibility-based aggregation leads to worse performance than the default design. We attribute this degradation to errors in the predicted depth maps, which can make the visibility estimates unreliable during multi-view fusion. In contrast, the softmax-based aggregation provides more robust feature integration across views, despite not explicitly modeling visibility. This observation is also consistent with the trend reported in BKinD-3D.

\begin{table}[]
\footnotesize
% \vspace{-12pt}
\setlength{\tabcolsep}{2pt}
\caption{Comparison of inference-time computational cost on an NVIDIA A6000 GPU.}\label{table:rebuttal_computation}
% \vspace{-12pt}
\resizebox{1\columnwidth}{!}{
\begin{tabular}{lccccc}
\toprule
Model            & Framework  & FLOPs (GFLOPs) & VRAM (GB) & Time/Iter (ms) & Throughput (iter/s) \\ \hline
(2D) StableKeypoints  & PyTorch    & 880.19         & 9.01           & 241.32         & 4.14                \\
(3D) KeypointNet      & TensorFlow & 700.83         & 6.06           & 75.41          & 13.26               \\
(3D) BKinD-3D         & PyTorch    & 296.27         & 10.51          & 553.73         & 1.81                \\ \hline
Ours - diffusion & PyTorch    & 7,787.39       & 29.48          & 11,930.67      & 0.08                \\
Ours - keydiff3d  & PyTorch    & 291.10         & 11.69          & 200.79         & 4.98                \\ \bottomrule
\end{tabular}}
\vspace{-13pt}
\end{table}

\subsection{Inference Computation Cost}
We provide an analysis of the inference-time computational cost in \Tref{table:rebuttal_computation}, including comparisons with existing 2D and 3D keypoint discovery baselines. Although SV3D enables our framework to estimate 3D keypoints from a single image, it also introduces substantial computational overhead. We view this as a trade-off for a capability that, to the best of our knowledge, is not supported by prior methods, namely unsupervised 3D keypoint estimation for arbitrary object categories from a single image.

Importantly, the additional cost is dominated by the diffusion stage rather than the proposed keypoint estimation module itself. As shown in \Tref{table:rebuttal_computation}, our KeyDiff3D module has computational cost comparable to existing 3D keypoint methods, while the diffusion model accounts for most of the extra FLOPs, memory usage, and latency required for monocular 3D keypoint estimation. Moreover, our framework uses diffusion features obtained from partial denoising, which is more efficient than relying on fully denoised generated images.

\section{Additional implementation details}

\subsection{Additional loss}
For Stanford Dogs and CUB-200-2011 datasets, we introduce a
\textit{view-invariant consistency loss} to regularize 3D keypoint prediction
under the feature-space viewpoint perturbations.
Given an input image, we obtain 2D projections of the predicted 3D keypoints
$\mathbf{S}$.
We then apply a random 3D rotation to the intermediate 3D volume features,
predict a second set of keypoints, and project them to obtain
$\mathbf{S}_{\mathrm{rot}}$.
The loss encourages consistency between the two projections:
\[
\mathcal{L}_{\mathrm{vic}}
= \frac{1}{N} \sum_{i=1}^{N}
\left\| \mathbf{S}_i - \mathbf{S}_{\mathrm{rot}, i} \right\|_2^2,
\]
and is added to the total loss with a weighting factor of $\lambda_{vic} = 0.1$.

\subsection{Model architecture details}

\paragraph{Feature aggregator.}
The feature aggregator fuses multi-scale features extracted from different layers of the diffusion decoder. 
Prior to aggregation, all feature maps are bilinearly upsampled to a shared spatial resolution of $72 \times 72$.
Each feature map is processed by a bottleneck block consisting of three convolutional layers: a $1\!\times\!1$ projection layer, a $3\!\times\!3$ convolution, and a final $1\!\times\!1$ expansion layer.  
All layers are followed by GroupNorm (32 groups) and ReLU activation.  
In addition, the residual shortcut connection is applied through a $1\!\times\!1$ convolution.  
The output features are projected to a unified embedding dimension (default: 384) across all layers.
After processing all layers, the resulting features are aggregated via a learnable softmax-weighted sum. The weights are initialized uniformly and optimized during training. The final output is a fused feature map of shape $[B, 384, 72, 72]$.

\paragraph{Keypoint head.}
The keypoint head takes the aggregated 2D feature map as input and predicts multi-view keypoint features with $C''$ channels for each view. This head consists of a lightweight convolutional network with three layers. The first two layers use $3 \times 3$ convolutions followed by SiLU activation, progressively reducing the channel dimension (from 128 to 32). The final layer uses a $1 \times 1$ convolution to project the features to $N$ channels, followed by a Tanh activation to bound the output values. This module is shared across all views and produces a keypoint-specific feature map of a shape $[B, C'', H, W]$.

\paragraph{Volumetric decoder.}
The volumetric decoder predicts $N$ keypoint heatmaps over a voxel grid of resolution $72 \times 72 \times 72$ from an input 3D feature volume. To ensure computational efficiency and avoid excessive memory usage, we adopt a 3D CNN architecture inspired by V2V-PoseNet~\cite{moon2018v2v}, where a final 3D convolution layer outputs $N$ volumetric heatmaps, from which 3D keypoint coordinates are estimated via soft-argmax.

\paragraph{Reconstruction network.}
We employ an encoder-decoder network to reconstruct RGB images conditioned on edge and keypoint heatmaps. Unlike standard U-Net or hourglass networks, we do not use skip connections between encoder and decoder layers. Instead, the decoder directly incorporates the predicted edge and keypoint maps at each resolution level to guide structure-aware synthesis. The encoder consists of downsampling residual blocks, while the decoder performs upsampling and concatenates the corresponding edge and keypoint maps at each stage. The final output is passed through a $1 \times 1$ convolution followed by Tanh activation.

\subsection{Data preprocessing}
\paragraph{Foreground mask.}
To generate foreground masks for object-centric inputs, we leverage different segmentation models depending on the dataset.
For the Human3.6M dataset, we use SAM with the 2D keypoint locations provided as query points.  
For animal datasets, such as Stanford Dogs and CUB-200-2011, we use Grounded-SAM2, which allows text prompts as input.  
In this case, we use the class names provided in the dataset (\eg, ``dog'') as the prompt to extract object-specific masks.

\paragraph{Crop and align.}
To ensure that the foreground object is centered in the image, we perform cropping and alignment based on either keypoints or segmentation masks.  
For Human3.6M, we compute a bounding box that tightly encloses all projected 2D keypoints, and shift it to the center of the image.  
For other datasets, such as Stanford Dogs, we instead compute a bounding box from the foreground mask.  
All images are then resized to $576 \times 576$, which matches the input resolution of the pretrained SV3D model.  
During evaluation on Human3.6M, we map the predicted results back to the original image space using the inverse of the affine transformation matrix obtained from the OpenCV's \texttt{warpAffine} function.

\section{Video results}
Please refer to the accompanying file \textit{demo.mp4} for more comprehensive descriptions of our framework, 3D keypoint results across diverse object categories, and animation results.

\begin{figure*}[t]
\begin{center}
\includegraphics[width=1\linewidth]{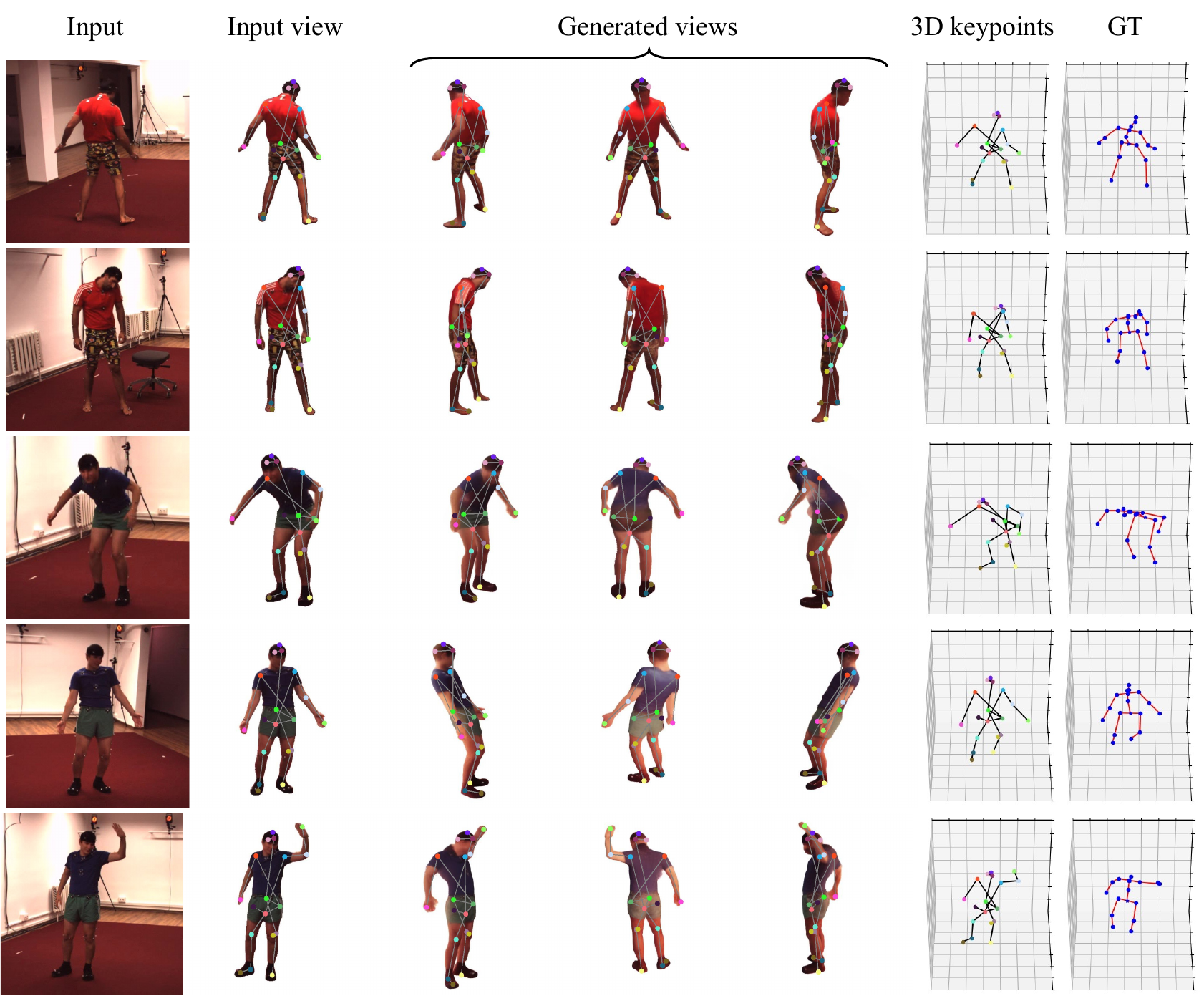}
\end{center}
   \caption{Additional results on Human3.6M dataset. }
   \label{fig:supple_hm36}
   \vspace{0pt}
\end{figure*}
\begin{figure*}[t]
\begin{center}
\includegraphics[width=1\linewidth]{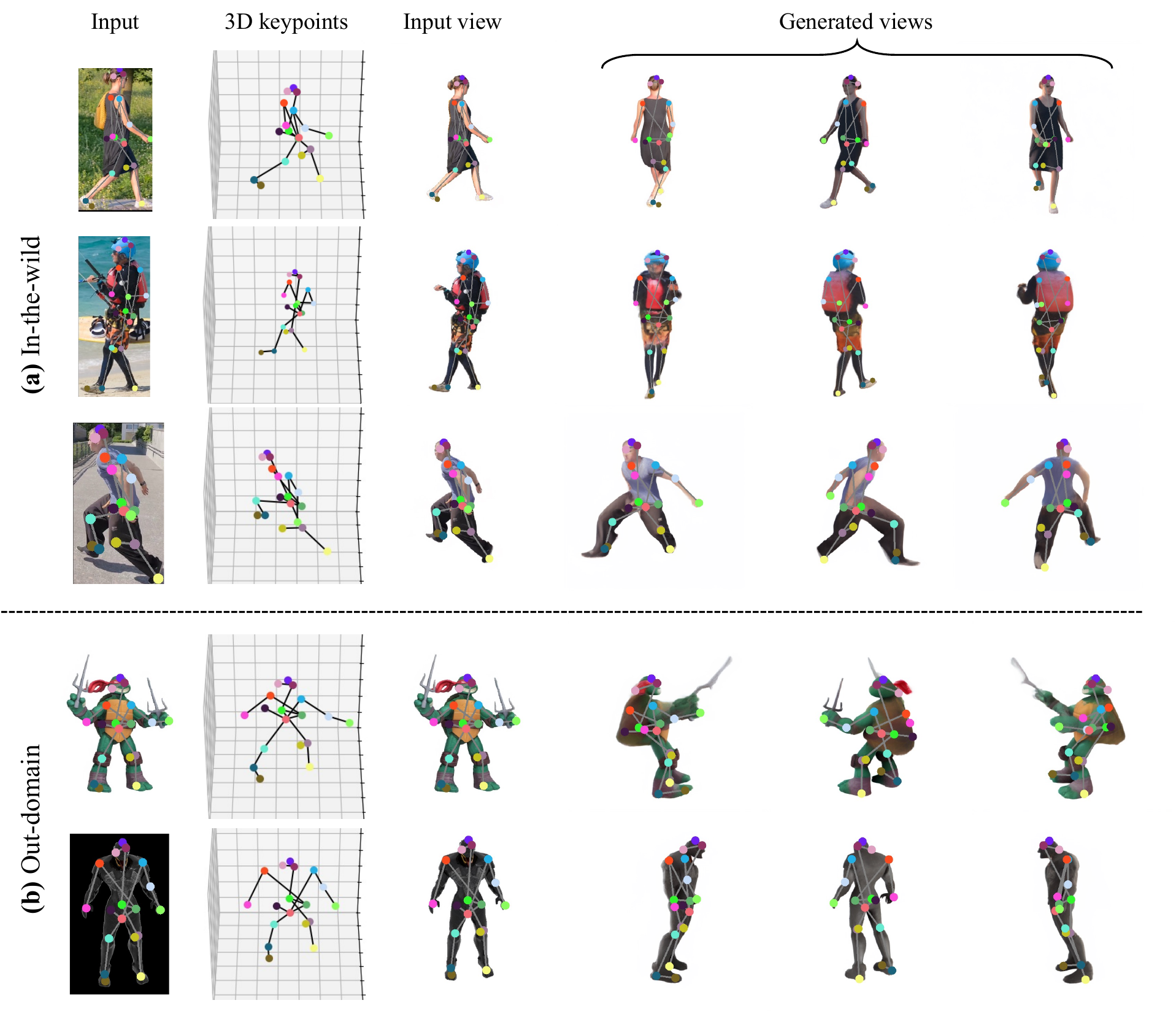}
\end{center}
   \caption{Results of our model trained with the Human3.6M dataset, which is tested on (a) in-the-wild and (b) out-of-domain images.}
   \label{fig:supple_wild}
   \vspace{0pt}
\end{figure*}
\begin{figure*}[t]
\begin{center}
\includegraphics[width=1\linewidth]{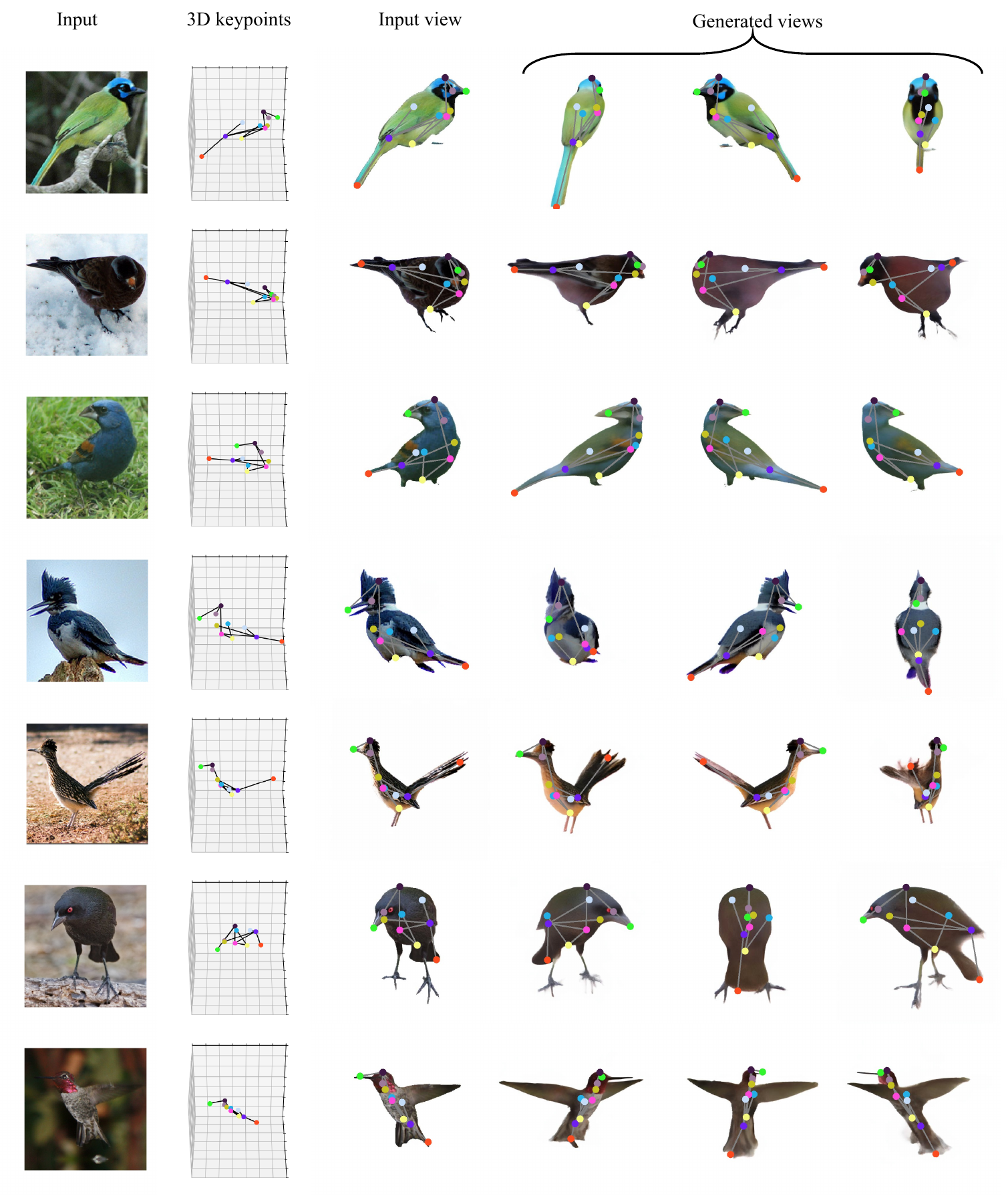}
\end{center}
   \caption{Additional results on CUB-200-2011 dataset (CUB-align) }
   \label{fig:supple_cub_align}
   \vspace{0pt}
\end{figure*}
\begin{figure*}[]
\vspace{-10pt}
\begin{center}
\includegraphics[width=0.95\linewidth]{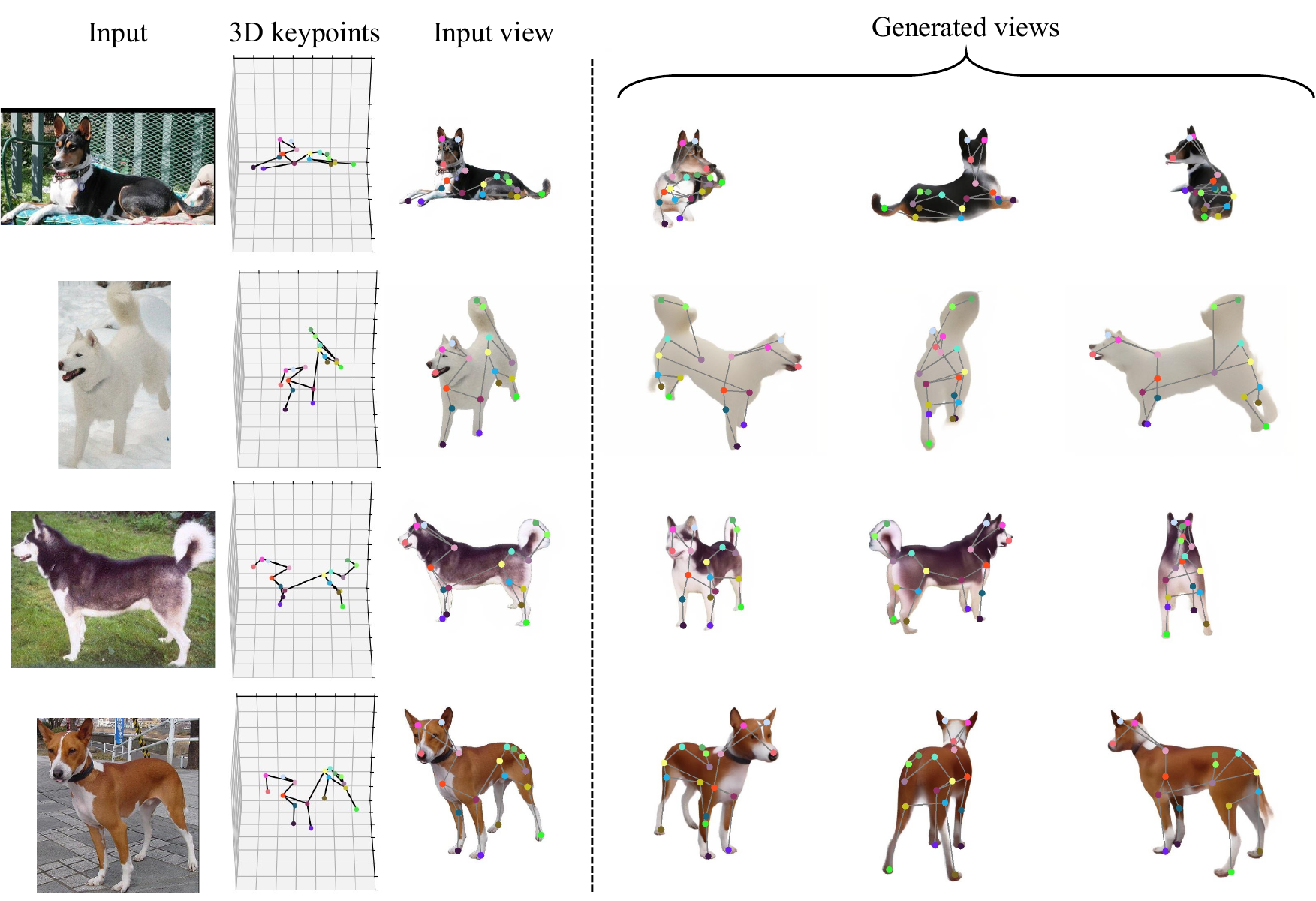}
\end{center}
   \vspace{-10pt}
   \caption{Additional results on Stanford Dogs dataset. }
   \label{fig:supple_dogs}
   \vspace{-12pt}
\end{figure*}
\begin{figure*}[t]
\begin{center}
\vspace{-3pt}
\includegraphics[width=0.95\linewidth]{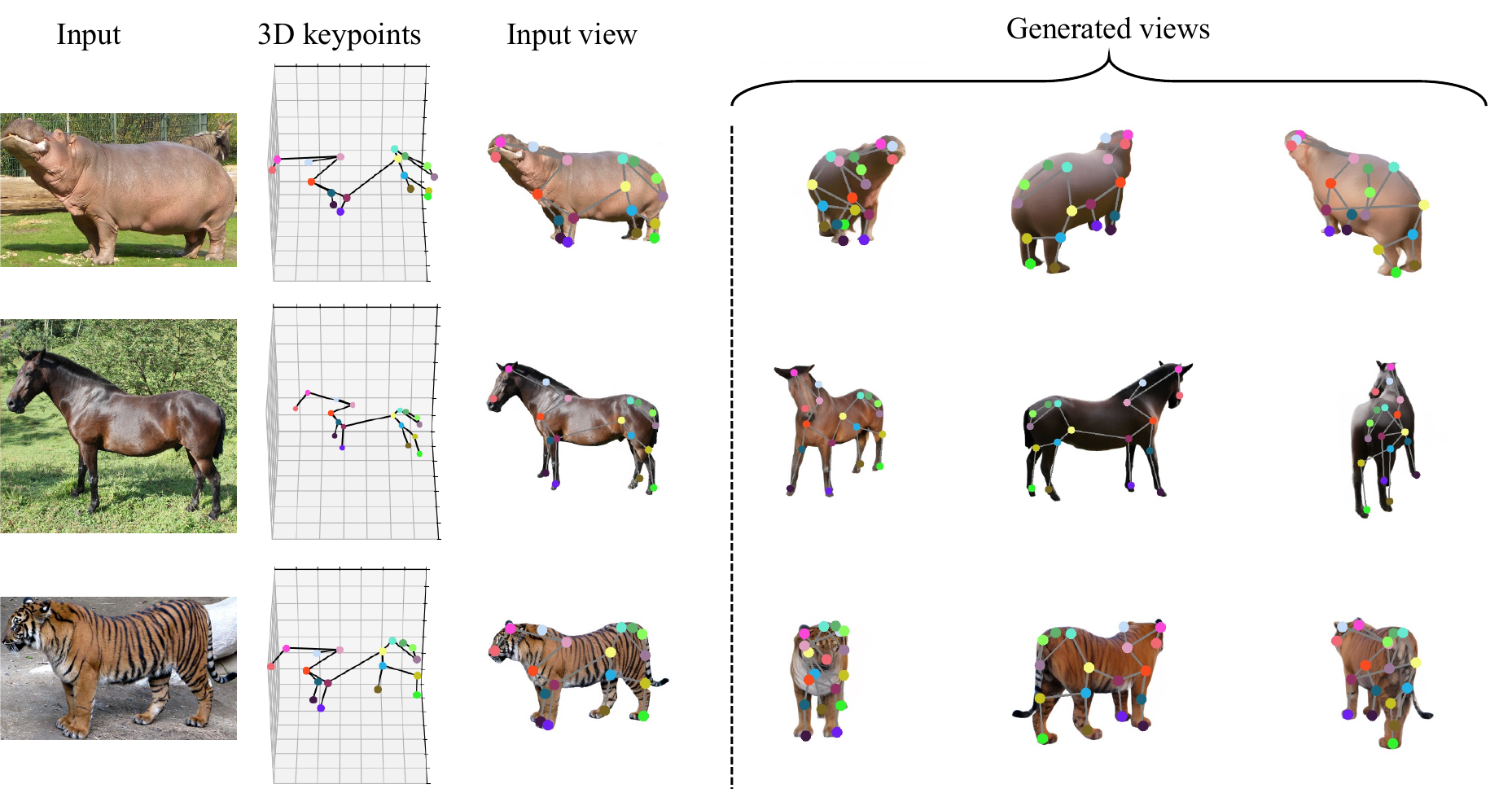}
\end{center}
\vspace{-10pt}
   \caption{Additional results on the AP-10K dataset using a model trained on Stanford Dogs.}
   
   \label{fig:supple_animal}
   \vspace{-10pt}
\end{figure*}

\end{document}